%% file: AAAI26-DiLA.tex
\documentclass[letterpaper]{article} 
\usepackage{aaai2026}  
\usepackage{times}  
\usepackage{helvet}  
\usepackage{courier}  
\usepackage[hyphens]{url}  
\usepackage{graphicx} 
\usepackage{natbib}  
\usepackage{caption} 
\graphicspath{{./figs/}{../}}
\newcommand{\minisection}[1]{\vspace{.08in}\noindent{\textbf{#1}}.}

\usepackage{graphicx}
\usepackage{subfigure}
\usepackage{multirow}

\usepackage{algorithm}
\usepackage{algorithmic}
\usepackage{amsmath}
\usepackage{cleveref}
\usepackage{amssymb}
\usepackage{pgfplots}
\usepgfplotslibrary{groupplots}
\usepackage{booktabs}

\usepackage{tikz}
\usetikzlibrary{positioning,calc,fit,decorations.pathmorphing,shapes.geometric,shapes.gates.logic.US,calc}
\usepackage{tikz-3dplot}

\definecolor{myblue}{RGB}{71,162,241}      
\definecolor{myred}{RGB}{228,46,36}        
\definecolor{myorange}{RGB}{239,194,149}   
\definecolor{mygray}{RGB}{218,218,218}
\definecolor{mypurple}{RGB}{195,155,200}

\renewcommand{\vec}[1]{\boldsymbol{#1}}

\date{}

\title{DiLA: Enhancing LLM Tool Learning with Differential Logic Layer}

\author{
Yu Zhang\textsuperscript{\rm 1},
Hui-Ling Zhen\textsuperscript{\rm 2},
Zehua Pei\textsuperscript{\rm 1},\\
Yingzhao Lian\textsuperscript{\rm 2},
Lihao Yin\textsuperscript{\rm 2},
Mingxuan Yuan\textsuperscript{\rm 2},
Bei Yu\textsuperscript{\rm 1},
}
\affiliations {
\textsuperscript{\rm 1}The Chinese University of Hong Kong,\\
\textsuperscript{\rm 2}Noah’s Ark Lab, Huawei,\\
}

\begin{document}

\maketitle
\thispagestyle{plain}
\pagestyle{plain}

\input{doc/abstract}
\input{doc/intro}

\input{doc/background}
\input{doc/motivation}
\input{doc/method}

\input{doc/exp}

\input{doc/conclu}

\clearpage
{
\bibliography{ref/Top, ref/reference}
}

\input{doc/appendix}

\end{document}

%% file: doc/abstract.tex
\begin{abstract}

Logical reasoning remains a significant challenge for large language models (LLMs), particularly in tasks involving complex constraint satisfaction such as Boolean satisfiability (SAT) and graph coloring. Existing approaches—ranging from pure prompting-based reasoning to solver-aided frameworks—either suffer from unfaithful reasoning or face scalability bottlenecks due to exponential search spaces in symbolic solvers. 
In this paper, we present \textbf{DiLA} (Differential Logic Layer-Aided Language Modeling), a novel framework that integrates a differentiable logic layer into LLMs to jointly leverage linguistic understanding and gradient-based logical refinement. DiLA first translates natural language problems into SAT specifications and generates an initial LLM-informed variable assignment, then iteratively refines it through a logic layer implementing differentiable MaxSAT optimization. This synergy enables efficient reasoning grounded in formal logic while maintaining semantic awareness.
Comprehensive experiments across logical deduction, SAT, and graph coloring benchmarks demonstrate that DiLA achieves 100\% accuracy with up to 65$\times$ runtime speedup over solver-aided methods such as SATLM. On industrial-scale benchmarks where state-of-the-art solvers (Z3, Kissat) fail within 10,000 seconds, DiLA successfully converges in under 300 seconds, illustrating its robustness in large and highly constrained settings. 
Furthermore, on the Natural Language Constraint Reasoning benchmark, DiLA reaches 87\% end-to-end success rate, outperforming both SATLM and pure LLM baselines by large margins. 

\end{abstract}

%% file: doc/intro.tex
\section{Introduction}

Recently, a significant research thrust has been on leveraging large language models (LLMs) for reasoning and planning, with numerous efforts aimed at augmenting their reasoning capabilities. These endeavors include text-based temporal reasoning~\citep{llm-temporal-reasoning}, logic feedback-enhanced alignment methods~\citep{enhancing-rlhf}, and prompt-based methods such as the chain of thoughts (CoT)~\citep{cot} or a simple directive like ``Let's think step by step''~\citep{step-by-step}. To further bolster the reasoning ability of LLMs, an AlphaZero-like tree-search learning framework has been introduced. This framework demonstrates how tree-search, coupled with a learned value function, can guide LLMs' decoding and enhance their reasoning abilities~\citep{alphazero-llm}. Despite the strides made by LLMs in achieving human-like reasoning abilities, they still encounter challenges when confronted with complex plannings~\citep{valmeekam2022large}. Occasionally, LLMs exhibit unfaithful reasoning, leading to derived conclusions that do not consistently follow the previously generated reasoning chain in practical applications~\citep{logic-llm}. 


To this end, a growing trend in this area involves augmenting LLMs with access to external engines, i.e., utilizing LLMs to first parse natural language logical questions into symbolic representations and subsequently employ external solvers to generate answers based on these representations. To enhance parsing accuracy, LoGiPT~\citep{llm-logical-solvers} has been proposed to directly emulate the reasoning processes and mitigate the parsing errors by learning to strict adherence to solver syntax and grammar. It is fine-tuned on a constructed instruction-tuning dataset derived from revealing and refining the invisible reasoning process of deductive solvers. An alternative approach for improving reasoning capabilities involves SATLM, a new satisfiability-aided modeling approach~\citep{SatLM}, in which an LLM is used to generate a declarative task specification with few-shot prompting and offload the actual reasoning task to an off-the-shelf theorem prover, i.e., Z3 solver~\citep{z3}. This method has demonstrated superiority over program-aided Language Models~\citep{pal} by achieving a $23\%$ improvement on the GSM arithmetic reasoning dataset, establishing a new state-of-the-art.  

\begin{figure}[t]
    \centering
    \includegraphics[width = \linewidth]{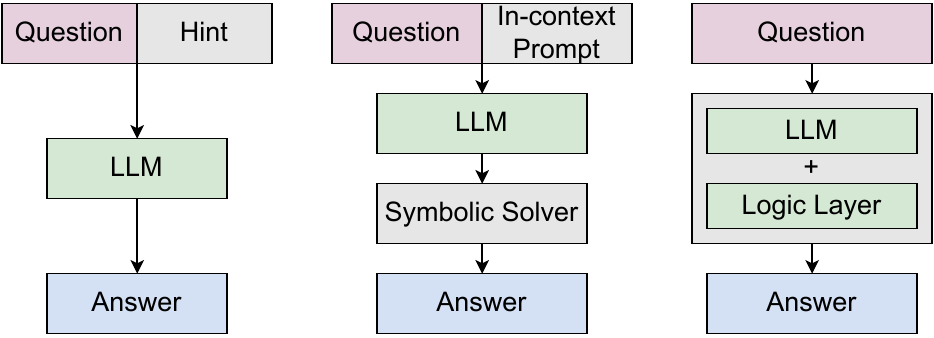}
    \caption{Illustration of CoT (left), solver-aided approach (middle), and our logic layer-aided language modeling approach (right).}
    \label{fig:block}
\end{figure}

Despite their impressive performance on various benchmark tests, SATLM~\citep{SatLM} and its solver-augmented successors can not deal with reasoning and planning problems in practical scenarios. Particularly, these real problems often state a set of premises and complex constraints and require a sophisticated search process to find the optimal solution, which is still challenging even for modern solvers. More specifically, even the state-of-the-art symbolic solvers, such as Z3\citep{z3} and Kissat\citep{kissat}, encounter significant bottlenecks when addressing formal verification problems such as electronic circuit verification involving tree/cyclic circuit structures\citep{satformer} and job-shop scheduling problems~\citep{schedule} with complex graph connections.
For instance, the Kissat solver requires days or even weeks to process a vanilla circuit verification problem with arithmetic circuit modules including multipliers or multiply-add circuits, yet fails to produce a solution. This highlights the limitations of solver-augmented LLM approaches which solely rely on symbolic solvers to tackle logical reasoning problems in reality.

In this paper, we propose DiLA, a novel tool-learning approach for Large Language Models, designed to enhance their logical reasoning capabilities through the integration of an additional logic layer. Unlike existing methods that either rely on in-context prompting for step-by-step reasoning (see Figure~\ref{fig:block}(left)) or entirely offload reasoning to external solvers (see Figure~\ref{fig:block}(middle)), our approach offers a third option: enhancing LLMs' reasoning ability by incorporating a differential logic layer (see Figure~\ref{fig:block}(right)). Specifically, DiLA leverages the LLM to parse and comprehend the problem description, generate an initial solution based on its language understanding, and then iteratively refine this solution through forward and backward passes of a network layer that embeds first-order logic constraints into its architecture. In this way, DiLA overcomes the limitations of traditional solvers by directly performing reasoning within the framework of layer-augmented LLMs. Our contributions are summarized as follows:

\begin{itemize}
    \item We introduce a novel tool-learning approach for LLMs, DiLA, which synergistically integrates a differential logic layer into LLMs, effectively bridging the gap between natural language understanding and symbolic reasoning capabilities.
    \item Leveraging SAT encoding as a bridge, DiLA successfully translates natural language reasoning problems into satisfiability problems, enabling it to tackle a range of reasoning problems, such as SAT and GCP.
    \item We propose a novel neural network layer, termed the logic layer, which differentiates symbolic problems and iteratively searches for solutions through forward and backward propagation of a network layer, thereby circumventing the limitations of off-the-shelf solvers.
\end{itemize} 

We evaluate the performance on three constraint satisfaction problems: logical deduction, Boolean satisfiability, and graph coloring. Our analysis yields two key findings: firstly, on simple problem instances, DiLA boosts the inference accuracy of LLMs to 100\% and consistently outperforms solver-aided approaches with improved runtime. Secondly, for real-world problems that current solvers struggle with, DiLA showcases robustness and remarkable efficiency in handling these complex test cases, thereby opening up opportunities for further real-world applications.


%% file: doc/background.tex
\section{Preliminary and Related Works}

\subsection{Preliminary}

In general, Boolean formulae are represented in Conjunctive Normal Form (CNF) as a conjunction
of clauses, where a clause is a disjunction of literals and a literal denotes either a variable or its negation.
Each variable can be assigned a logic value, either $0$ or $1$.
Any general Boolean problem can be represented as a CNF formula model.
An SAT solver either finds an assignment such that CNF is satisfied or proves that no such assignment exists, i.e., UNSAT.
Modern SAT solvers are based on the conflict-driven-clause-learning (CDCL) algorithm, and they work as a basic engine for many applications. SAT plays an important
role in data mining applications, e.g., Maximal Frequent Subgraph
Mining~\cite{liu2022satmargin}, Graph Coloring~\cite{velev2007exploiting}, Sequence Mining~\cite{jabbour2013boolean}.


\begin{figure*}[t]
    \centering
    \includegraphics[width = .84\linewidth]{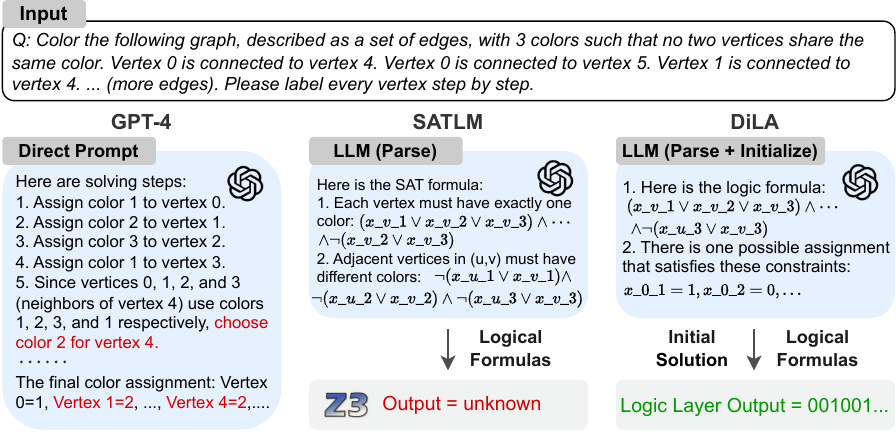}
    \caption{Exemplar comparison of solving graph coloring problems by different approaches. Direct prompts by GPT-4 make errors when generating the color assignment step by step; SATLM, based on the Z3 symbolic solver, cannot solve parsed constraints and outputs unknown (both errors are highlighted in red). In contrast, DiLA can generate the correct answer by combining the strengths of LLMs and the differential logic layer.}
    \label{fig:compare}
\end{figure*}

\subsection{Related Works}

\minisection{Logic Reasoning in LLMs}
Prior approaches to NL-based reasoning with LLMs can be broadly categorized into two groups. One is \emph{in-context learning} approaches that design special prompts to elicit LLMs’ step-by-step reasoning capabilities. Typical methods include chain-of-thought prompting~\citep{cot} that generates a sequence of reasoning steps before the final answer and the least-to-most prompting~\citep{least2most} that breaks the problem down into simpler components that can be solved sequentially. Both the above approaches perform reasoning directly over natural language (NL), providing greater flexibility than symbolic-based reasoning. However, the intrinsic complexity and ambiguity of NL also bring undesired issues such as unfaithful reasoning and hallucinations. The other is \emph{tool-augmented} approaches that only require LLMs to parse the problem specification out of the language description accurately and leverage off-the-shelf automated tools to derive the final answer, as exemplified by SATLM~\citep{SatLM} and LOGIC-LM~\citep{logic-llm}. The tool-augmented approaches guarantee the correctness of the answer with respect to the parsed specification and avoid planning errors in the solving process. However, the performance of such tool-augmented methods highly relies on external tools and can not deal with many real problems due to the deficiency of backbone solvers.

\minisection{Neural Symbolic Reasoning}
Deep neural networks have made remarkable progress in various domains, but their ability to handle logical reasoning tasks remains uncertain. One potential approach is to encode logical constraints that are essential for certain inference
tasks symbolically, making them available to the model either during training or inference. 
SATNet~\citep{wang2019satnet}, for instance, introduces a differentiable MaxSAT solver layer into its network architecture to learn the logical structure of problems through data-driven learning. Their approximation is based on a coordinate descent approach for solving the semidefinite program (SDP) relaxation of the MaxSAT problem. SATNet does not assume that the logical structure of the problem is given, and instead attempts to learn it. 
While SATNet achieves high accuracy in solving Sudoku problems, it cannot be generalized to regular SAT problems as it relies on a specific number of accurate assignments as inputs and is only applicable to solving logic problems with constant constraints after training. 
In the pursuit of neuro-symbolic learning, SMTLayer~\citep{smtlayer} incorporates a satisfiability modulo theories (SMT) solver into a DNN layer, thereby embedding problem-specific theories into DNN architectures through data-driven training. Compared to SATNet, SMTLayer demonstrates superior accuracy in logical reasoning tasks and a reduced need for training data. However, SMTLayer still requires a certain amount of training data and relies on off-the-shelf SMT solvers for reasoning, making it unsuitable for industrial problem-solving. Similar to the aforementioned neuro-symbolic approach, DiLA also aims to encode logical constraints in the network layer, making them available to the model during forward and backward passes.

%% file: doc/motivation.tex
\section{Motivation}

This paper explores the potential of language understanding and logical reasoning capabilities in LLMs. Traditionally, the prevailing approaches have either relied solely on LLMs for step-by-step reasoning~\citep{cot, least2most} or offloaded reasoning tasks to off-the-shelf solvers~\citep{linc, SatLM}. However, we propose a third approach, as these two extremes either underutilize or over-rely on LLMs during reasoning. Specifically, DiLA leverages the powerful understanding abilities of LLMs to extract logical formulas and generate possible solutions. By utilizing LLMs, such as GPT-4, as a solution generation engine, we can produce an initial solution based on input semantic constraints, laying the groundwork for further refinement.



The second motivation stems from an analysis of current solver applications in real-world scenarios, characterized by two distinctive features:
(1) The large scale of reasoning problems in reality, leading to rapid degradation in the performance of heuristic-based solvers due to the exponential expansion of the search space;
(2) The formidable challenge presented by the complex structure of logic formulas, often requiring weeks or even months to resolve using current solvers.
Both characteristics significantly limit the effectiveness of a solver-augmented LLM in logic reasoning.

Consequently, our objective is to identify a synergistic approach that combines the strengths of LLMs and a differential logic layer, thereby circumventing the limitations of traditional solvers. This approach leverages the LLM's capacity to comprehend logical formulas while concurrently utilizing the logic layer's refinement abilities to achieve accurate solutions. \Cref{fig:compare} illustrates a comparison between CoT, SATLM, and our proposed DiLA. The LLM alone may introduce logical flaws during step-by-step inference, such as assigning the same color to vertex 1 and vertex 4 despite their edge connection, and SATLM may struggle with complex reasoning problems due to its backbone solver's limitations. In contrast, DiLA produces accurate answers through the collaboration of the LLM and the differential logic layer. Specifically, DiLA first uses the LLM to parse a natural language input into logic constraints and generate an initial solution based on its semantic understanding, then employs the differential logic layer to refine this initial solution. 

\begin{figure*}[tb!]
    \centering
    \includegraphics[width =.88\linewidth]{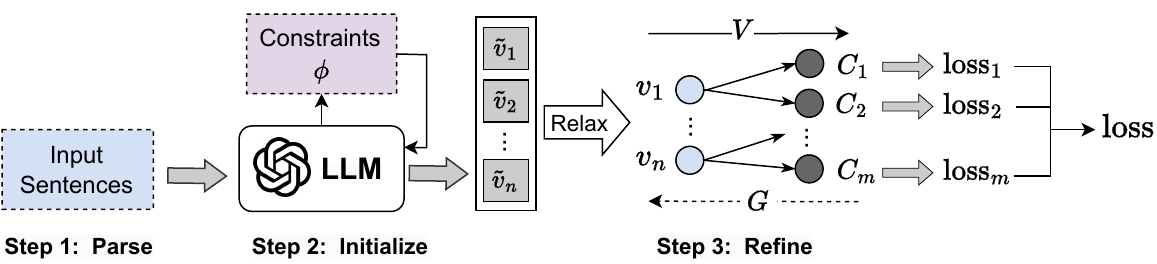}
    \caption{The overall flow of DiLA. }
    \label{fig:dila}
\end{figure*}


%% file: doc/method.tex
\section{Differential Logic Layer-Aided LLMs}

\subsection{Overview}

In this section, we present DiLA, which augments LLM with the ability of logical reasoning by incorporating a differential logic layer. More specifically, DiLA addresses the challenge of using LLMs to tackle canonical reasoning tasks expressed in natural language. These tasks typically involve presenting a set of premises and constraints, prompting questions that necessitate intricate deductive reasoning over the provided inputs, which remains a formidable challenge even for contemporary LLMs~\citep{valmeekam2022large}.


The general procedure for solving natural language reasoning tasks with DiLA can be conceptualized in three distinct steps: parsing, initialization, and refinement (as illustrated in \Cref{fig:dila}). Given a natural language input that describes both the propositional constraints $\phi$ and the question $Q$, we first parse this input into a SAT specification using LLMs (\textit{step 1}), thereby obtaining a formal description of the constraints and variables. Next, we leverage the LLM's natural language understanding to generate an initial variable assignment (\textit{step 2}). Since this initial solution may only partially satisfy the propositional constraints, DiLA iteratively refines it through a differential logic layer that encodes all of the logical formulas (\textit{step 3}), ultimately yielding a more accurate solution.

\subsection{Problem Formulator and Initialization}\label{sec:initial}

Intuitively, LLMs may struggle with directly solving complex reasoning problems.
However, they have demonstrated a notable ability to comprehend textual inputs and translate them into formal programs, such as mathematical equations~\citep{math} or satisfiability modulo~\citep{SatLM}.
Notably, the SAT problem can serve as a versatile intermediate step for solving a broad range of constraint satisfaction problems, provided they can be expressed using Boolean variables. Specifically, problem instances from NP-complete domains, such as Graph Coloring and Set Cover, can be seamlessly encoded into SAT problem specifications, thereby allowing for efficient solutions via SAT algorithms~\citep{graphcolor, satlayout}. Therefore, we harness SAT encoding as a general bridge to tackle these constraint satisfaction problems in practical settings.

Specifically, given a problem description in natural language, DiLA prompts an LLM with detailed instructions to generate the SAT specification, which includes a set of premises and constraints. Typically, the SAT specification here involves conjunctive normal forms (CNFs), denoted as $\phi(v_1, \dots, v_n)$, which is a conjunction of clauses (constraints) $C$. Formally, rules are written in the conjunctive form of clauses $C_1 \wedge C_2 \cdots \wedge C_m $, which each $C_j$ is a constraint. The total rule is satisfied if and only if all of the clauses $C_1, C_2, \dots, C_m$ are simultaneously True. Each clause represents a disjunction of literals, where a literal is either a propositional variable $v_i$ or its complement $\neg v_i$, for example, $v_1 \vee \neg v_2$. In other words, if at least one literal in a clause is True, this clause would also be True.
Variables can be assigned logic values, either $1$ or $-1$, representing True or False, respectively\footnote{In other works, they may claim the logic value of each literal is $0$ or $1$. It should be noted that the two claims are equal under simple mathematical transformations.}.


Aside from problem formulation, leveraging an LLM can be a valuable strategy to generate an initial solution. We observe that, after generating the SAT specification, we can prompt LLMs, like GPT-4~\citep{gpt4} or Llama-3~\citep{llama3}, to produce a potential solution or a set of possible solutions. This can be achieved by framing the problem as a natural language query, such as ``What is the logical solution based on the premises?'' or ``Can you provide a possible answer from these given constraints?''. The LLM's response can then serve as a starting point for further refinement and validation, allowing us to build upon its output and iteratively improve the solution through logical analysis and reasoning. By harnessing the LLM's ability to understand semantic constraints, we can tap into its potential to facilitate the initial solution-finding process and accelerate solving progress towards a well-reasoned answer.

\subsection{From SAT to Differential MaxSAT}\label{sec:maxsat}

Traditionally, a SAT solver, such as Z3 and Kissat, is leveraged to determine a satisfying assignment for the given constraint formula $\phi$. Contemporary SAT solvers are founded on the CDCL algorithm, which excels in its ability to learn from conflicts and use that conflict knowledge to prune the branch-and-bound search space more effectively. However, existing CDCL-based SAT solvers still suffer from exponential searching space and are unable to correct errors through a learning-from-mistakes system, resulting in an infinite loop in solving complex SAT problems~\citep{satformer}.

In this study, when an NL reasoning problem is fed into DiLA, our goal is to determine its solution that can satisfy all logical constraints. To achieve this, we might incorporate a solver as an additional logic reasoner for the LLM, as in SATLM~\citep{SatLM} or LOGIC-LM~\citep{logic-llm}. However, the exponential search complexity inherent in heuristic-based SAT solvers poses a significant challenge, limiting their effectiveness when dealing with complex real-world problems. Therefore, a key issue is how to design an efficient solver surrogate that can both be seamlessly integrated into LLMs and efficiently address logical reasoning problems.

The maximum satisfiability (MaxSAT) problem serves as the optimization counterpart to the SAT problem, aiming to maximize the number of satisfied clauses. Indeed, if a solution to the MaxSAT problem can satisfy all the clauses, the variable assignment can be used to constitute a valid solution for the original SAT problem. In the case of the SAT problem, each CNF is associated with a set of clauses (constraints), and each clause is defined on a subset of variables, signifying the variables' simultaneous legal assignments. Formally, each CNF $\phi(v_1, \dots, v_n)$ comprises $n$ binary variables, with each $v_i \in \{1, -1\}$ ($i \in {1,2,\dots,n}$) representing a boolean variable. The coefficients are represented as $\vec{c}_j \in \{-1,0,1\}^m$, where $c_{ij}$ denotes the sign of $v_i$ in clause $j \in {1,2,\dots,m}$. Consequently, we can establish a clause matrix $\vec{C} \in \{1,-1,0 \}^{m \times n}$, where each element $c_{ij}$ in $\vec{C}$ signifies the sign of variable $\Tilde{v}_i$ in clause $j$. Therefore, each SAT instance can be translated into a corresponding MaxSAT problem, wherein $\bigvee$ represents the logical ``or'' symbol,

\begin{equation} 
    \underset{\Tilde{\vec{v}} \in \{-1,1 \}^n}{\max } \sum_{j=1}^m \bigvee_{i=1}^n \textbf{1}\{c_{ij}  \Tilde{v}_i > 0 \}.\label{eq:1}
\end{equation}

We further formulate \Cref{eq:1} in its minimization, or \textit{unsatisfiability}, equilibrium as
\begin{equation}
    \underset{\Tilde{\vec{v}} \in \{-1,1 \}^n}{\min } \sum_{j=1}^m \bigwedge_{i=1}^n \textbf{1}\{c_{ij}  \Tilde{v}_i < 0 \},\label{eq:2}
\end{equation}
where $\bigwedge$ is the logical ``and'' symbol. Indeed, the objective value in \Cref{eq:2} is 0 if and only if a satisfiable solution can be found. Our goal is to establish a continuous upper bound, referred to as the ``loss'', for each clause to quantify its level of unsatisfiability. In essence, the loss takes an upper bound if the clause is unsatisfied, and by minimizing this loss, we can strive to push it closer to satisfaction. To make a purely quadratic loss function as in ~\citep{wang2019max}, we introduce $v_0 = 1$ and $s_{0j} = -1$ in \Cref{eq:2}. Therefore, the minimization problem in \Cref{eq:2} can be solved by transforming into a quadratic loss function as:
\begin{equation}
    \mathcal{L}_j = \frac{(\sum_{i=0}^n c_{ij}  \Tilde{v}_i)^2 - (m_j -1)^2}{4 m_j}, \quad \mathcal{L} = \sum_{j=1}^m \mathcal{L}_j , \label{eq:3}
\end{equation}
where $\mathcal{L}_j$ is the loss value of $j$-th clause, $\mathcal{L}$ is the loss value of all clauses, and $m_j$ is the number of literals in clause $j$, e.g., 3 for the Max3SAT problem. The loss function in \Cref{eq:3} is essentially a quadratic loss that takes the upper bound when no literal in clause $j$ is satisfied. In other words, it captures the extent of unsatisfiability for a given clause by penalizing solutions that fail to satisfy any of its literals. Specifically, for any value of $m_j$, it can be easily verified that this quantity is equal to +1 if no literal is satisfied, and 0 or less if at least one literal is True. 

Take a simple SAT problem with clauses $(v_1 \vee v_2) \wedge v_1$ as an illustrating example.
Based on \Cref{eq:3}, the loss function for this SAT problem is
\begin{equation*}
    \text{loss} = \text{loss}_1 + \text{loss}_2 = \frac{(v_1+v_2-v_0)^2 - 1}{8} + \frac{(v_1 - v_0)^2}{4}.
\end{equation*}

Now, the MaxSAT solving is equivalent to finding an assignment vector $\Tilde{\vec{v}} \in \{-1,1 \}^n$ that minimizes loss in \Cref{eq:3}. By relaxing each discrete variable $\Tilde{\vec{v}}_i$ to a continuous variable $\vec{v}_i \in \mathbb{R}$, the quadratic loss function becomes
\begin{equation}
    \mathcal{L}_j = \frac{\| \vec{V} c_j \|^2 - (m_j -1)^2}{4 m_j},\label{eq:4}
\end{equation}
which is essentially a convex minimization problem. Therefore, leveraging gradient descent to solve this minimization problem, the gradient computation involves differentiating the loss function in \Cref{eq:4} with respect to $v_i$. Define this gradient as $\vec{g_i}$, we have

\begin{equation}
    \vec{g}_i = \vec{V} \vec{S}^\top \vec{s}_i - \|\vec{s}_i \|^2 \vec{v}_i,\label{eq:5}
\end{equation}
where $\vec{S} = [\vec{c}_0, \vec{c}_1, \dots, \vec{c}_n ] \text{diag}(1/\sqrt{4 m_j}) \in \mathbb{R}^{m \times (n+1)}$ and $s_i \in \mathbb{R}^{(n+1)}$ is the $i$-th vector in $\vec{S}$.

\subsection{Differential Logic Layer}\label{sec:layer}

We envision the logic layer being used primarily at the top of LLMs, embedding logical formulas $\phi$ produced by the backbone LLM, taking LLM-initialized variable assignments as inputs, and producing outputs that are consistent with $\phi$. Specifically, we draw an analogy between the fully connected (FC) layer and the variable-clause graph (VCG), where the weight for positive literals in the clause is 1 (solid arrow) and negative literals in the clause is -1 (dashed arrow). Specifically, we map each variable to an input neuron in an FC layer, each clause to an output neuron, and the coefficients to the weights in the linear transformation with a zero bias vector. While this analogy holds, there are two crucial differences between traditional FC layers and our proposed logic layer. Firstly, the logic layer has no unknown parameters, whereas FC layers require data-driven training to learn their weights. Secondly, each clause in the logic layer is only partially connected to variables, in contrast to fully connected layers, where all input neurons are connected to all output neurons.

Supposing that the current solution of MaxSAT in \Cref{eq:1} is given as $\phi^\prime = \phi(v_1,v_2,\dots,v_n)$, and it is easy to check whether this solution, i.e., variable assignment, can satisfy the original SAT problem $\phi$, which is essentially the \emph{forward pass} of our logic layer. A trivial case is that if all the clauses (i.e., constraints) are satisfied by the assignments $\mathcal{C}(v_1, v_2, \dots, v_n)$, then these assignments constitute a valid solution for the original SAT problem $\phi$. In most scenarios where $\phi^\prime$ satisfies only a subset of the clauses, we define the unsatisfied clauses as $\bar{\phi^\prime}$, a subset of all clauses, and denote the indices of the variables involved in $\bar{\phi^\prime}$ as $\bar{I}$. Intuitively, the variables in $\bar{I}$ are likely to be the source of conflicts, so we select the variable with the largest absolute gradient from the candidate set $\bar{I}$ and update its value during the backward pass, thereby pushing $\phi^\prime$ towards satisfying more constraints. We now elaborate on the forward and backward pass of our proposed logic layer, providing a detailed explanation of its operation.

\minisection{Forward Pass}
The forward pass algorithm is outlined in Algorithm~\ref{algorithm:forward}. In the forward pass, the inputs consist of relaxed variable assignment at $k$-th iteration. Subsequently, the layer transforms these inputs by extracting the sign of the variables, thereby casting them to Boolean values. The layer then assesses the satisfiability of $\phi^\prime$ (line 2). If the current variable assignment satisfies $\phi^\prime$, the logic layer outputs $y^k$ as True, indicating that $\phi$ is satisfied and a feasible solution for the given CNF has been identified. Conversely, if $\phi$ cannot be satisfied, the logic layer outputs $y^k$ as False, prompting the initiation of the backward pass to update the variable assignment. 




\begin{algorithm}[H]
    \caption{The forward pass of logic layer}
    \begin{algorithmic}[1]
    \REQUIRE{Solution $\vec{v}^k \in \mathbb{R}^n$ at $k$-th epoch.}
    \ENSURE{$y^k$, final solution $\vec{v}^*$.}
        \STATE $\tilde{\vec{v}}^k \gets [\vec{v}^k_i>0: i=1,\dots,n ]$;
        \STATE $\phi^\prime \gets \phi(\tilde{v}^k_1, \dots, \tilde{v}^k_n)$;
        \IF{$\phi^\prime$ is satisfiable}
        \STATE $y^k \gets$ True;
        \STATE $\vec{v}^* \gets \tilde{\vec{v}}^k$ ;
        \ELSE
        \STATE $y^k \gets$ False;
        \ENDIF
    \end{algorithmic}
    \label{algorithm:forward}
\end{algorithm}

\begin{table*}[tb!]
    \centering
    \caption{
        Accuracy and runtime (s) of CoT, including GPT-4, Llama-3, and DeepSeek-R1, SATLM, and DiLA on LogicalDeduction,
        Boolean Satisfiability and 3-Coloring datasets. Llama-3 SC denotes the Llama-3 model with Self-Consistency~\cite{wang2022self}.
    }
    \label{table:easy}
    \resizebox{0.8\linewidth}{!}
    {
        \begin{tabular}{|c|c|c|c|c|c|cc|cc|}
            \hline
            \multirow{2}{*}{Problem} & \multirow{2}{*}{\#Variables} & \multicolumn{1}{c|}{GPT-4} & \multicolumn{1}{c|}{Llama-3}&  \multicolumn{1}{c|}{Llama-3 SC} & \multicolumn{1}{c|}{DeepSeek-R1}& \multicolumn{2}{c|}{SATLM} & \multicolumn{2}{c|}{DiLA} \\ 
                \cline{3-10} 
                &    & Acc (\%) & Acc (\%) & Acc (\%) & Acc (\%) & Acc (\%)   & Time (s) & Acc (\%) & Time (s) \\ \hline \hline
                \begin{tabular}[c]{@{}c@{}}Logical\\ Deduction\end{tabular}
                    & 3,5,7& 76 & 81 & 83 & 97 &  100  & 0.01 &  100  &  0.01 \\ \hline
                    \multirow{5}{*}{SAT} & 20 & 12 & 17 & 17 & 39 &  100   & 0.05 &  100  &  0.01 \\ 
                    & 50& 5 & 0 & 0 & 7 &  100   & 0.12 &  100  &  0.01 \\ 
                    & 100& 0  & 0 & 0 & 0 &  100   & 0.17 &  100  &  0.03 \\ 
                    & 200& 0 & 0 & 0 & 0 &   100  & 0.99 &  100  &  0.04 \\ 
                    & 250& 0 & 0 & 0 & 0 &  100   & 5.24 &  100  & 0.08  \\ \hline
                    \multirow{5}{*}{\begin{tabular}[c]{@{}c@{}}Graph\\ Coloring\end{tabular}}   
                    & 10 & 20 & 22 & 22 & 34 &  100   & 0.05 &  100  &  0.01 \\ 
                    & 50& 7 & 10 & 10 & 15 &  100   & 0.19 &  100  &  0.02 \\ 
                    & 100& 0 & 0 & 0 & 0 &  100   & 0.58 &  100  &  0.04 \\ 
                    & 150& 0 & 0 & 0 & 0 &   100  & 2.64 &  100  & 0.15  \\ 
                    & 200& 0 & 0 & 0 & 0 &  100  & 5.70 &  100  &  0.24 \\ \hline
        \end{tabular}
        }
\end{table*}

\minisection{Backward Pass}
The backward pass is responsible for computing the gradients of the layer inputs and derives updates to variables that steer towards satisfying the constraints $\phi$. A crucial aspect of the backward pass is identifying the input variables that contribute most to the unsatisfiability of the constraint formulas. It is well-established that variables in $\bar{I}$ form the unsatisfiable subset and are, therefore, more likely to be sources of conflict. Conversely, variables not present in $\bar{I}$ can have their gradients set to zero, as their absence in the conflict clauses provides no evidence regarding the correctness or incorrectness of their values. Inspired by the stochastic local search (SLS) algorithm, commonly used in constraint satisfaction problems~\citep{chu2023nuwls}, we select the variable with the largest absolute gradient from the candidate set $\bar{I}$ and update its value at each iteration. However, our logic layer diverges from SLS in that it employs a ``differential'' variable selection mechanism during backpropagation, whereas SLS relies on meta-heuristics.

\begin{algorithm}[htb!]
    \caption{The backward pass of logic layer}
    \begin{algorithmic}[1]
    \REQUIRE{$\vec{v}^k \in \mathbb{R}^n$ from forward pass, $y^k$ from forward pass, learning rate $\lambda$.}
    \ENSURE{Gradient $\vec{g}^k$ of $\vec{v}^k$, updated assignment $\vec{v}^{k+1}$.}
    \STATE $\vec{g}^k \gets \emph{0}$;
        \IF{$y^k$ is False}
        \STATE $\tilde{\vec{v}}^k \gets [\vec{v}^k_i>0: i=1,\dots,n ]$;
        \STATE $\phi^\prime \gets \phi(\tilde{v}^k_1, \dots, \tilde{v}^k_n)$;
        \STATE $\bar{I} \gets \{ i \in [1,n] | v_i \in \bar{\phi^\prime}\}$;
        \FOR{$ i \in \bar{I}$}
        \STATE $g_i \gets  \partial_{\vec{v}^k_i} \mathcal{L}$;
        \ENDFOR
            \IF{$\exists \; g_i \neq 0$}
                \STATE $g^k_i \gets \arg \max_{i \in \bar{I}}  \|g_i \|$;
            \ELSE
                \STATE $v_i :=$ a random variable in a falsified clause;
                \STATE $g^k_i := \text{sign}(v_i)$;
            \ENDIF
        \ENDIF
        \STATE Update $\vec{v}^{k+1} \gets \vec{v}^k - \lambda \vec{g}^k $;
    \end{algorithmic}
    \label{algorithm:backward}
\end{algorithm}

Algorithm~\ref{algorithm:backward} illustrates our backward pass.
The backward pass begins by initializing the gradient to zero for all variables (line 1).
If $y^k$ is false, indicating the presence of unsatisfied clauses, we obtain the set of variables $\bar{I}$ that are present in the falsified clauses $\bar{\phi^\prime}$.
Once we have obtained the candidate set $\bar{I}$ (line 5), we proceed to select the best variable from this set based on its gradient.
Specifically, we compute the gradient as in \Cref{eq:5} for each variable in the candidate set (line 7).
Then, the logic layer selects a variable and updates its value based on two situations:
(1) If there exists a variable with a non-zero gradient (i.e., $g_i \neq 0$), the variable with the largest absolute $g_i$ would be selected (line 10);
(2) If there is no variable satisfying the above condition, indicating that the search is stuck in a local optimum, we randomly select a variable from a falsified clause (line 12) and artificially assign a gradient that would change its sign after gradient descent (line 13). More discussions are given in Appendix.

%% file: doc/exp.tex
\section{Experiments}

In this section, we present an empirical evaluation of DiLA on solving logical reasoning problems expressed by natural language. Particularly, we test our approach in satisfiable datasets, as tackling unsatisfiable (UNSAT) certification presents a distinct and separate challenge. Although DiLA is not directly applicable to UNSAT problems, we present an extension that identifies potential unsatisfiable cores in Appendix\ref{appendix:unsat}. Many real-world applications (e.g., resource allocation, route planning) naturally admit solutions when given sufficient flexibility. For instance, in graph coloring, while a graph may be UNSAT for 3 colors, it often becomes satisfiable when more colors are permitted. For satisfiable instances, LLMs need to determine the CNF instance's satisfiability and output the variable assignments that genuinely satisfy the input CNF instances. 


\begin{table*}
    \begin{minipage}[p]{0.6\textwidth}
\centering
\footnotesize
\caption{Runtime (s) of Z3, Kissat, and DiLA on hard problems. DiLA successfully handles these hard cases where canonical solvers struggle.}
\resizebox{\linewidth}{!}
        {
\begin{tabular}{|c|ccc|ccc|}
\hline
Problem   & Test case & \#Variables & \#Clauses & Z3 & Kissat & DiLA  \\ \hline \hline
\multirow{4}{*}{SAT}   & rbsat & 1150    & 84314 & \textgreater{}10,000 & \textgreater{}10,000 & 98.76 \\
 & sgen3 & 260  & 884   & \textgreater{}10,000 & \textgreater{}10,000 & 11.83 \\
 & Schur & 756  & 28445 & \textgreater{}10,000 & \textgreater{}10,000 & 23.81 \\
 & SCPC  & 900  & 41714 & \textgreater{}10,000 & \textgreater{}10,000 & 20.23  \\ \hline
\multirow{4}{*}{\begin{tabular}[c]{@{}c@{}}Graph\\ Coloring\end{tabular}} & g125.17   & 125*17  & 68397 & \textgreater{}10,000 & \textgreater{}10,000 & 29.57 \\
 & g125.18   & 125*18  & 72413 & \textgreater{}10,000 & \textgreater{}10,000 & 31.14  \\
 & g250.15   & 250*15  & 237715    & \textgreater{}10,000 & \textgreater{}10,000 & 29.68  \\
 & g250.29   & 250*29  & 461872    & \textgreater{}10,000 & \textgreater{}10,000 &  30.07 \\ \hline
\end{tabular}
}
\label{tab:hard}
\end{minipage}
  \begin{minipage}[p]{0.38\textwidth} 
    \centering 
    \input{figs/compare}
    \captionof{figure}{Comparing DiLA with and without LLM initialization.}
    \label{fig:runtime}
  \end{minipage}
\end{table*}

\subsection{Experiment Setup}

\minisection{Tasks}
We conduct experiments on three fundamental reasoning tasks: the logical deduction problem, Boolean Satisfiability, and Graph Coloring problems. The logic deduction problems are mostly about deducing the order of a sequence of objects from a minimal set of conditions. Here we utilize the LogicalDeduction dataset from the BigBench~\citep{bigbench} collaborative benchmark. For SAT problems, we utilize open-source benchmark instances~\footnote{https://www.cs.ubc.ca/\textasciitilde hoos/SATLIB/benchm.html} with 20 to 250 variables, focusing on finding variable assignments that satisfy all constraints. For Graph Coloring problems, we randomly generated 100 3-coloring instances with vertex counts ranging from 10 to 200, following the approach in ~\citep{graphcolor} (see Appendix\ref{appendix:data_generation} for details), aiming to color graph vertices so that no two adjacent vertices share the same color.
Furthermore, we also incorporate several complex SAT cases in SAT Competition 2023 and open-source graph coloring problems with a large number of edges to illustrate the robustness of our proposed DiLA. These instances that originate from practical industrial problems are known to be satisfiable, reflecting real-world scenarios where solutions are expected to exist. They frequently pose challenges for modern SAT solvers. Notably, current solvers often enter into an infinite loop when handling these test cases due to their complex structure, requiring a prohibitively long time to solve.

\minisection{Baselines}
We conducted a comparative analysis between DiLA and several baselines, including the CoT-based methods with GPT-4~\citep{gpt4} and Llama-3 (Llama-3-70B-Instruct)~\citep{llama3}, Self-Consistency~\cite{wang2022self} enhanced Llama-3, DeepSeek-R1~\cite{guo2025deepseek}, and the solver-augmented method, SATLM~\citep{SatLM}. In particular, SATLM~\citep{SatLM} employs GPT-4 to parse problem specifications and offloads the logical reasoning task to the symbolic solvers, i.e., Z3 solver~\citep{z3}, which serves as the state-of-the-art tool-learning method for logic reasoning. 

\minisection{Setup}
We implement a prototype of our proposed DiLA using Pytorch~\citep{pytorch}, leveraging GPT-4 as the backbone LLM model. Notably, the logic layer within DiLA has no training parameters and can adapt to various problem types expressed in Boolean variables. We employ the SGD optimizer with a learning rate of $2 \times 10^{-1}$ in DiLA, facilitating the effective updating of selected variables. Furthermore, we use a temperature of 0 for LLMs, consistent with the SATLM approach. We set a time limit of 10,000 seconds for both the solvers and the logic layer. All experiments are performed on 8 NVIDIA V100 (32GB), with LLM-involved stages dominating memory usage. The logic layer of DiLA can be performed on a single GPU for benchmarks with hundreds of thousands of variables/clauses, due to sparse connections and low-bit layer weights.

\subsection{Main Results}\label{sec:main_result}

We report the accuracy of DiLA and baselines in Table~\ref{table:easy}. Accuracy is evaluated based on whether the LLM can output a correct answer that satisfies all constraints. We evaluate LLMs over 100 instances in each domain. In addition to accuracy, we also report the solving runtime for SATLM and DiLA, both of which leverage GPT-4 as a backbone LLM to perform language understanding. 

Analysis of Table~\ref{table:easy} reveals that both solver-augmented SATLM and logic layer-aided DiLA achieve 100\% accuracy on these simple benchmarks, owing to the precise logic parsing and accurate solving. In contrast, standalone LLMs, including GPT-4, Llama-3, and DeepSeek-R1, cannot handle reasoning problems with over 100 variables with prompting-based methods, i.e., chain-of-thought and self-consistency. Specifically, while self-consistency prompting improves performance on constrained logical deduction tasks, its benefits diminish for relatively large testcases. The DeepThink reasoning flow in DeepSeek-R1 enables notable improvements on small benchmarks. However, challenges remain for problems exceeding 50 variables/nodes. In contrast, for all test instances, DiLA exhibits faster performance than SATLM in the solving process, especially for relatively large cases with over 200 variables. The runtime speedup can be up to 65.5$\times$ when dealing with SAT problems with 250 variables and 23.8$\times$ when tackling 3-coloring problems with 200 nodes. Overall, both the solver-augmented LLM and our proposed DiLA can successfully address these simple artificial test cases, with the runtime speedup highlighting the efficiency of DiLA in logic reasoning.

\subsection{Generalization to Industrial Benchmarking}\label{sec:hard_problem}

To investigate the performance boundaries and demonstrate the differentiability of DiLA from solver-aided methods, we evaluate it on a set of hard problems derived from the real world. We compare DiLA against two distinct and powerful solvers: Z3 and Kissat. For all methods, we set a stringent time limit of 10,000 seconds per instance. All experiments were conducted on the same hardware configuration as previously described.

The experimental results, presented in Table~\ref{tab:hard}, demonstrate that DiLA successfully solves these challenging cases within a reasonable runtime, whereas both Z3 and Kissat fail to produce valid results within the time limit. As indicated in Table~\ref{tab:hard}, these difficult constraint satisfaction instances typically exhibit large clause-to-variable (CV) ratios (m/n). For example, the test case ``rbsat'' consists of only 1150 variables, but the total number of clauses amounts to 84,314, resulting in an extremely high CV ratio of 73.32. This suggests that there is likely only one viable solution for these problems. If a traditional solver's initial search path deviates significantly from the correct path, the CDCL framework's inability to rectify errors through a learning-from-mistakes system can lead to an endless loop. In contrast, our proposed DiLA, initialized by LLMs and guided by differentiation of the loss function, enables us to first reach a partially satisfied solution and then progressively update it using an efficient searching strategy, rather than completely failing if stuck. In general, the evaluation results on industrial cases demonstrate the high efficiency of our proposed DiLA.


\minisection{Ablation study of LLM initialization} The comparison results, shown in \Cref{fig:runtime}, indicate that the initial solutions generated by backbone LLMs can serve as an excellent starting point. Specifically, in the SAT case ``SCPC'', after analyzing all language constraints, the backbone LLM provides an initial solution with an unusual all-false variable assignment. We discovered that over 99\% of the final satisfying variable assignments should indeed be set to 0, indicating that a significant proportion of variables require no further updates. 
In contrast, traditional solvers rely on random initialization and need to explore all possible assignments for each variable before reaching the final solution, resulting in exponential search spaces and reduced efficiency. 

\subsection{Natural Language Reasoning }

While previous sections demonstrated DiLA's effectiveness on standard benchmarks, we now evaluate its performance on a more challenging task: solving constraint satisfaction problems described entirely in natural language, without any structured inputs. This evaluation directly addresses the concern about the parsing capability of the backbone LLM and demonstrates DiLA's end-to-end reasoning capability from raw text to validated solutions.

We construct a new benchmark, \textbf{Natural Language Constraint Reasoning (NLCR)}, specifically designed to test the synergy between language understanding and logical reasoning. Each problem in NLCR requires three key capabilities: (1) extracting implicit constraints from unstructured natural language text, (2) translating semantic constraints into formal SAT specifications, and (3) finding satisfying variable assignments that meet all requirements.

The benchmark comprises 100 problems across four realistic domains: scheduling problems (30 instances) involving temporal and resource constraints, resource allocation (25 instances) with capacity and assignment rules, planning problems (25 instances) requiring sequential decision-making, and configuration problems (20 instances) with dependency constraints. An instance from NLCR benchmark is illustrated in Appendix\ref{sec:nlr}. 

\subsubsection{Comparative System Evaluation}

We compare three representative approaches that span from pure neural to hybrid neuro-symbolic methods: (1) \textbf{Pure LLM} leveraging GPT-4 with chain-of-thought prompting to directly generate solutions, (2) \textbf{SATLM} combining LLM parsing with the Z3 symbolic solver for formal reasoning, and (3) \textbf{DiLA} integrating LLM parsing and initialization with our differential logic layer for iterative refinement.

Table~\ref{tab:nlcr_overall} presents the overall performance across all 100 NLCR problems. DiLA achieves 87\% end-to-end success rate, outperforming SATLM (76\%) and Pure LLM (51\%) by significant margins. Notably, DiLA also demonstrates superior efficiency with an average solving time of 4.5 seconds, faster than SATLM's 8.7 seconds despite handling more complex reasoning internally. The parsing accuracy of DiLA (92\%) exceeds both baselines, suggesting that the iterative refinement capability of the logic layer provides implicit feedback that improves constraint extraction quality.

\begin{table}[t]
\centering
\caption{Performance comparison on NLCR benchmark}
\label{tab:nlcr_overall}
\small
\setlength{\tabcolsep}{3pt}
\begin{tabular}{@{}lcccc@{}}
\toprule
\textbf{System} & \textbf{Parse} & \textbf{Solution} & \textbf{E2E} & \textbf{Time} \\
 & \textbf{Acc} & \textbf{Quality} & \textbf{Succ.} & \textbf{(s)} \\
\midrule
Pure LLM & 78\% & 65\% & 51\% & 3.2 \\
SATLM & 85\% & 90\%$^*$ & 76\% & 8.7 \\
\textbf{DiLA} & \textbf{92\%} & \textbf{95\%} & \textbf{87\%} & \textbf{4.5} \\
\bottomrule
\end{tabular}
\end{table}

Breaking down performance by problem category in Table~\ref{tab:nlcr_category}, we observe that DiLA's advantages are most pronounced in scheduling tasks (+17\% over SATLM), where temporal reasoning and resource constraints interact in complex ways. The improvement is consistent across resource allocation (+12\%) and planning problems (+12\%), while configuration problems show comparable performance. This pattern suggests that DiLA's strength lies in handling problems with intricate constraint interactions that challenge both pure LLM reasoning and traditional solver heuristics.

\begin{table}[ht]
\centering
\caption{End-to-end success rate by problem category.}
\label{tab:nlcr_category}
\small
\begin{tabular}{@{}lcccc@{}}
\toprule
\textbf{Category} & \textbf{LLM} & \textbf{SATLM} & \textbf{DiLA} & \textbf{Gain} \\
\midrule
Scheduling & 47\% & 73\% & \textbf{90\%} & +17\% \\
Resource & 52\% & 76\% & \textbf{88\%} & +12\% \\
Planning & 48\% & 72\% & \textbf{84\%} & +12\% \\
Config. & 60\% & 85\% & \textbf{85\%} & 0\% \\
\midrule
\textbf{Overall} & 51\% & 76\% & \textbf{87\%} & \textbf{+11\%} \\
\bottomrule
\end{tabular}
\end{table}

\subsubsection{Quantifying Semantic Understanding}

Beyond overall performance metrics, we investigate whether DiLA's LLM component provides value beyond constraint parsing through semantic understanding. We design 20 problems with explicit semantic biases where understanding the problem structure should guide initialization. For instance, in a star-shaped graph coloring problem with one central node connected to 10 peripheral nodes, the semantic insight is that the central node should receive a ``rare'' color since it must differ from all peripherals.

We define a Semantic Understanding Score (SUS) as the fraction of variables that receive semantically meaningful initial assignments:
\begin{equation}
\text{SUS} = \frac{\text{\# semantically correct initial assignments}}{\text{\# total variables}}
\end{equation}

Table~\ref{tab:semantic_score} demonstrates that LLM initialization achieves a SUS of 0.89, correctly handling 18 out of 20 semantic-rich problems. This substantially exceeds random initialization (SUS = 0.18, 2/20 correct) and heuristic-based initialization (SUS = 0.45, 9/20 correct). The high semantic understanding translates directly into faster convergence, as the logic layer starts from a more informed initial state that already captures key problem structure.

\begin{table}[t]
\centering
\caption{Semantic Understanding Score across 20 problems.}
\label{tab:semantic_score}
\small
\begin{tabular}{@{}lcc@{}}
\toprule
\textbf{Method} & \textbf{SUS Score} & \textbf{Correct} \\
\midrule
Random & 0.18 & 2/20 \\
Heuristic & 0.45 & 9/20 \\
\textbf{LLM (DiLA)} & \textbf{0.89} & \textbf{18/20} \\
\bottomrule
\end{tabular}
\end{table}




%% file: figs/compare.tex
\begin{tikzpicture}[scale=1]
\tikzstyle{every node}=[font=\small]
\begin{groupplot}[
    group style={
        group name=my plots,
        group size=1 by 1,
    },
    width=\linewidth,
    height=.6\linewidth,
    ybar=12pt,
    enlargelimits=0.25,
    enlarge y limits=0,
    legend style={
        at={(0.55,0.99)},
        draw=none,
        anchor=north west,
        legend columns=1,
        legend style={font=\tiny},
    },
    area legend,
    y label style={at={(0.26,0.5)}},
    symbolic x coords={sgen3, g250.29, SCPC},
    xtick=data,
    ytick pos=left,
    nodes near coords,
    nodes near coords align={vertical},
]
\nextgroupplot[
    ybar = 6pt,
    bar width=12pt,
    ylabel={Runtime$\!~(s)$},
    y label style={at={(0.07,0.5)}},
    ymin=0.0,
    ymax=315.0,
    ytick={0.0, 300.0, 150},
]
    \addplot[fill=mygray,   area legend] coordinates {(sgen3,107.33) (g250.29,269.65) (SCPC,137.22)}; 
    \addplot[fill=mypurple, area legend] coordinates {(sgen3,11.83) (g250.29,30.07) (SCPC,20.23)}; 
\legend{w/o.~LLM, w.~LLM}
\end{groupplot}
\end{tikzpicture}

%% file: doc/conclu.tex
\vspace{-4pt}
\section{Conclusion}

In this work, we introduce a pioneering method named differential logic layer-aided language modeling (DiLA). Starting with an NL reasoning problem, DiLA first uses an LLM to cast it into a SAT problem and generate a possible solution based on its language understanding, and then progressively refines this solution within a logic layer. In this way, we harness the potential of the language-understanding ability of LLMs and sidestep the limitations of off-the-shelf solvers. Extensive experiments on two reasoning tasks demonstrate the superior efficiency of our approach over state-of-the-art solver-augmented LLMs. On large-scale industrial verification and cryptographic problems, DiLA retains its efficiency advantage where state-of-the-art solvers (Z3, Kissat) fail to converge within their time budgets, underscoring its robustness and scalability. Beyond traditional SAT formulations, DiLA exhibits strong synergy between language understanding and logical reasoning in natural language constraint reasoning tasks, achieving 87\% end-to-end success rate on our newly proposed NLCR benchmark. The results suggest that DiLA achieves new state-of-the-art performance in symbolic logical reasoning tasks, paving the way for more applications of LLMs in practical reasoning settings. 





%% file: doc/appendix.tex
\clearpage
\appendix







\section{More Evaluations on Natural Language Reasoning} \label{sec:nlr}

To test the limits of each approach, we evaluate performance on 10 large-scale NLCR instances with 20--30 entities, 50--80 constraints, and 500--1000 SAT variables after encoding. These problems represent realistic complexity levels encountered in industrial applications.

Figure~\ref{fig:nlcr_example} presents a representative scheduling problem that illustrates the complexity of natural language constraint specification in our benchmark.

\begin{figure}[htbp]
\centering
\fbox{
\begin{minipage}{0.95\columnwidth}
\small
\textbf{Problem ID: NLCR-Schedule-015}

\vspace{4pt}
\textit{Description:} A company has 5 meeting rooms (R1--R5) and needs to schedule 8 meetings (M1--M8) during a single day. Each meeting lasts 1 hour, and the working day spans 9 AM to 6 PM (9 time slots).

\vspace{4pt}
\textit{Constraints:}

\begin{enumerate}
    \item Meeting M1 must occur before M3.
    \item M2 and M4 cannot overlap.
    \item M5 requires room R1 (morning only).
    \item M6 and M7 must be adjacent.
    \item R3 cannot be used $>$2 consecutive hours.
    \item M8 must be the last meeting.
    \item No room hosts multiple meetings simultaneously.
\end{enumerate}

\vspace{4pt}
\textit{Task:} Provide a feasible schedule assigning each meeting to a room and time slot.
\end{minipage}
}
\caption{Example problem from the NLCR benchmark requiring natural language understanding and constraint reasoning.}
\label{fig:nlcr_example}
\end{figure}

Table~\ref{tab:nlcr_large} reveals stark performance differences across systems. Pure LLM fails on all instances (0\% success rate), unable to maintain reasoning coherence at this scale due to the depth and breadth of constraint interactions. SATLM achieves 40\% success rate, which can be decomposed into its constituent success factors: 60\% parsing accuracy multiplied by 67\% solver success rate (4 out of 6 correctly parsed instances solved within timeout). The solver timeouts occur primarily on instances with high clause-to-variable ratios where Z3's CDCL-based search struggles.

DiLA achieves 90\% success rate (9 out of 10 instances), demonstrating robustness at scale. This combines 92\% parsing accuracy with 98\% refinement success rate (all 9 correctly parsed instances solved within time limit). DiLA also provides substantial speedup over SATLM, averaging 5.0$\times$ faster solving time. On instances where SATLM times out (NLCR-L02, L04, L05, L07, L09, L10), DiLA completes in 45--89 seconds, demonstrating the efficiency advantages of gradient-based refinement over heuristic search for solver-hard instances.

\begin{table}[htbp]
\centering
\caption{Performance on large-scale NLCR problems with 20--30 entities and 50--80 constraints (timeout = 300s).}
\label{tab:nlcr_large}
\small
\resizebox{\linewidth}{!}{
\begin{tabular}{@{}lccccccc@{}}
\toprule
\textbf{Problem ID} & \textbf{\#Entities} & \textbf{\#Constraints} & \textbf{\#Variables} & \textbf{Pure LLM} & \textbf{SATLM} & \textbf{DiLA} & \textbf{Speedup} \\
\midrule
NLCR-L01 & 20 & 52 & 547 & Failed & 127s & \textbf{23s} & 5.5$\times$ \\
NLCR-L02 & 25 & 67 & 892 & Failed & Timeout & \textbf{45s} & $>$6.7$\times$ \\
NLCR-L03 & 22 & 58 & 634 & Failed & 203s & \textbf{31s} & 6.5$\times$ \\
NLCR-L04 & 28 & 73 & 1045 & Failed & Timeout & \textbf{67s} & $>$4.5$\times$ \\
NLCR-L05 & 30 & 81 & 1234 & Failed & Timeout & \textbf{89s} & $>$3.4$\times$ \\
NLCR-L06 & 23 & 61 & 701 & Failed & 178s & \textbf{38s} & 4.7$\times$ \\
NLCR-L07 & 27 & 70 & 967 & Failed & Timeout & \textbf{58s} & $>$5.2$\times$ \\
NLCR-L08 & 21 & 55 & 589 & Failed & 145s & \textbf{27s} & 5.4$\times$ \\
NLCR-L09 & 26 & 68 & 923 & Failed & Timeout & \textbf{71s} & $>$4.2$\times$ \\
NLCR-L10 & 29 & 77 & 1156 & Failed & Timeout & \textbf{82s} & $>$3.7$\times$ \\
\midrule
\textbf{Average} & 24.3 & 64.8 & 868.8 & \textbf{0\%} & \textbf{40\%} & \textbf{90\%} & \textbf{5.0$\times$} \\
\bottomrule
\end{tabular}
}
\end{table}

\subsection{Failure Mode Analysis}

To understand the limitations of each approach, we categorize all failures across the 100 NLCR problems in Table~\ref{tab:failure_modes}. The analysis reveals distinct bottlenecks for each system. Pure LLM exhibits a high rate of logical inconsistencies (27\%), where the generated solutions violate constraints despite the model's apparent understanding of individual requirements. This occurs because LLMs lack systematic constraint verification during step-by-step reasoning, leading to solutions that satisfy some constraints while inadvertently violating others.

SATLM's primary failure mode is solver timeout (17\%), occurring when the Z3 solver encounters complex formulas that exceed its heuristic search capabilities within the time limit. An additional 15\% of failures stem from parsing errors where the LLM misses implicit constraints or incorrectly formalizes natural language specifications. These parsing failures are particularly problematic for SATLM because the solver has no mechanism to recover from incomplete or incorrect constraint formulations.

DiLA achieves the lowest total failure rate (13\%) by mitigating both classes of failures. The logic layer's gradient-based refinement handles solver-hard instances that cause SATLM timeouts, reducing timeout failures to just 3\%. Meanwhile, the iterative nature of DiLA's reasoning process appears to improve parsing robustness, lowering parsing errors to 8\%. The virtual absence of partial solutions (0\%) demonstrates that when DiLA succeeds in parsing, the logic layer reliably finds complete satisfying assignments.

\begin{table}[htbp]
\centering
\caption{Failure mode distribution across 100 NLCR problems.}
\label{tab:failure_modes}
\small
\setlength{\tabcolsep}{4pt}
\begin{tabular}{@{}lccc@{}}
\toprule
\textbf{Failure Type} & \textbf{LLM} & \textbf{SATLM} & \textbf{DiLA} \\
\midrule
Parsing error & 22\% & 15\% & 8\% \\
Logical inconsistency & 27\% & 3\% & 2\% \\
Solver timeout & -- & 17\% & 3\% \\
Partial solution & -- & 8\% & 0\% \\
\midrule
\textbf{Total} & \textbf{49\%} & \textbf{24\%} & \textbf{13\%} \\
\bottomrule
\end{tabular}
\end{table}

The LLM component in DiLA provides value that extends beyond merely extracting explicit constraints from text. Figure~\ref{fig:implicit_constraint} illustrates this through a medical appointment scheduling problem involving both hard constraints (``Alice's appointment must be after Bob's'') and soft preferences (``Carol prefers mornings but can do afternoons if necessary'').

\begin{figure}[htbp]
\centering
\fbox{
\begin{minipage}{0.95\columnwidth}
\scriptsize
\textbf{Problem:} Schedule 3 medical appointments. Alice's appointment must be after Bob's. Carol prefers mornings but can do afternoons if necessary.

\vspace{4pt}
\textbf{SATLM Parsing:}
\begin{itemize}
    \item Constraint 1: $t_{\text{Alice}} > t_{\text{Bob}}$ \checkmark
    \item Constraint 2: cannot understand ``preference''
\end{itemize}

\vspace{4pt}
\textbf{DiLA with LLM Understanding:}
\begin{itemize}
    \item Constraint 1: $t_{\text{Alice}} > t_{\text{Bob}}$ (hard) \checkmark
    \item Constraint 2: Soft -- prefer $t_{\text{Carol}} \in \{9,10,11\}$ AM
\end{itemize}

\vspace{4pt}
\textit{Initial solution:}
Bob: 9 AM, Alice: 11 AM, Carol: 10 AM (honors preference)

\vspace{2pt}
Logic layer refines based on hard constraints while preserving soft preferences when possible.
\end{minipage}
}
\caption{Example showing DiLA's nuanced understanding of constraint priorities (hard vs. soft) beyond pure logical parsing.}
\label{fig:implicit_constraint}
\end{figure}

SATLM's parsing treats ``preference'' as non-binding and ignores it entirely, leading to solutions that may unnecessarily violate user preferences. In contrast, DiLA's LLM component recognizes the distinction between hard constraints and soft preferences, encoding this in the initial solution by assigning Carol to a morning slot (10 AM) while ensuring the hard constraint (Alice after Bob) is satisfied. The logic layer then refines this initialization while attempting to preserve the soft preference when compatible with hard constraints.

This example demonstrates three key capabilities of DiLA's LLM component: (1) nuanced understanding of constraint priorities (hard versus soft), (2) contextual reasoning that goes beyond pure logical parsing to capture user intent, and (3) generation of human-aligned solutions that respect implicit preferences embedded in natural language descriptions.

\subsection{Details of the SAT Specification}\label{appendix:specification}

To better leverage the parametric knowledge that LLMs have acquired from pretraining on vast amounts of language data, our approach uses a SAT specification as a bridge to encode a range of semantic constraints. Specifically, it translates general reasoning problems into formal constraints with Boolean variables. Below, we provide a prompt example of how to convert a classical graph 3-coloring problem into its SAT specification in Table~\ref{tab:formula_gen}.

\begin{table*}[ht]
\begin{tabular}{p{0.96\linewidth}}
\hline
\multicolumn{1}{c}{Example prompt of logic formula generation} \\ \hline
 \textbf{Question}: Color the following graph with 3 colors such that no two vertices share the same color. Vertex 0 is connected to vertex 4. Vertex 0 is connected to vertex 5. Vertex 1 is connected to vertex 4. $\cdots$ Vertex 8 is connected to vertex 5. Vertex 9 is connected to vertex 4. Vertex 9 is connected to vertex 5. There are a total of 10 vertices. What is the SAT formulation for this graph coloring problem?\\ \hline
 \textbf{LLM response}: The SAT (Satisfiability) formulation for the graph coloring problem involves encoding the problem as a Boolean satisfiability problem. Here's the SAT formulation for the graph coloring problem:\\
 \emph{Variables}: \\
 For each vertex v and each color c (Red, Blue, Green), we introduce a Boolean variable x\_vc that represents whether vertex v is assigned color c.In this case, we have 10 vertices and 3 colors, so we introduce 30 Boolean variables: x\_0\_Red, x\_0\_Blue, x\_0\_Green,..., x\_9\_Red, x\_9\_Blue, x\_9\_Green.\\
 \emph{Clauses}: \\
 We need to encode the following constraints:
    1. Each vertex must have exactly one color: For each vertex v, we add a clause that ensures exactly one color is assigned:\\
    $(x\_v\_Red \vee x\_v\_Blue \vee x\_v\_Green) \wedge \neg(x\_v\_Red \vee x\_v\_Blue) \wedge \neg(x\_v\_Red \vee x\_v\_Green)$ \\
    $ \wedge \neg(x\_v\_Blue \vee x\_v\_Green)$ \\
    This clause ensures that vertex v has at least one color (first part) and at most one color (second to fourth parts).\\
    2. Adjacent vertices must have different colors: For each edge (u, v) in the graph, we add a clause that ensures the vertices have different colors: \\
    $\neg(x\_u\_Red \vee x\_v\_Red) \wedge \neg(x\_u\_Blue \vee x\_v\_Blue) \wedge \neg(x\_u\_Green \vee x\_v\_Green)$\\
    This clause ensures that vertices u and v do not have the same color. \\
    \emph{Complete SAT formulation}: \\
    The complete SAT formulation consists of the conjunction of all the clauses:\\
    $(x\_0\_Red \vee x\_0\_Blue \vee x\_0\_Green) \wedge \neg(x\_0\_Red \vee x\_0\_Blue) \wedge \neg(x\_0\_Red \vee x\_0\_Green) \wedge \neg(x\_0\_Blue \vee x\_0\_Green) \wedge$\\
    $\cdots$ \\
    $(x\_9\_Red \vee x\_9\_Blue \vee x\_9\_Green) \wedge \neg(x\_9\_Red \vee x\_9\_Blue) \wedge \neg(x\_9\_Red \vee x\_9\_Green) \wedge \neg(x\_9\_Blue \vee x\_9\_Green) \wedge$ \\
    $\neg(x\_0\_Red \vee x\_4\_Red) \wedge \neg(x\_0\_Blue \vee x\_4\_Blue) \wedge \neg(x\_0\_Green \vee x\_4\_Green) \wedge$ \\
    $\cdots$ \\
    $\neg(x\_9\_Red \vee x\_5\_Red) \wedge \neg(x\_9\_Blue \vee x\_5\_Blue) \wedge \neg(x\_9\_Green \vee x\_5\_Green)$
 \\ \hline
\end{tabular}
\caption{Prompt example of converting a classical graph 3-coloring problem into its SAT specification.}
\label{tab:formula_gen}
\end{table*}

We can then leverage the transformed logic formula to construct the proposed logic layer. Specifically, for large graphs with many nodes and edges, we ask the LLM to generate Python code to help transform the natural language description into a SAT specification, ensuring the process is both quick and accurate.

After parsing logic formulas from the problem description, we can directly ask LLM to generate a possible solution according to its language understanding. The possible question prompt can be ``What is the logical solution based on the premises?'' or ``Can you provide a valid answer from these given constraints?''. 

\section{Discussions on UNSAT} \label{appendix:unsat}

In DiLA, we indeed find the solution for the MaxSAT problem in \Cref{eq:1}, and we denote the maximum satisfiable set of clauses as $\phi$. For an UNSAT problem, the complementary set of $\phi$, i.e., the remaining unsatisfied clauses $\bar{\phi}$, must include at least one clause from minimally unsatisfiable subformulas, i.e., UNSAT core. Therefore, based on the MaxSAT results from DiLA, we can design a procedure to iteratively find the UNSAT core:
    \begin{enumerate}
        \item \textbf{MaxSAT Solution}: DiLA first finds the maximal satisfiable subset of clauses $\phi$ for the problem.
        \item \textbf{Source of Conflict}: The complementary unsatisfied clauses $\bar{\phi}$ must contain at least one clause from UNSAT cores.
        \item \textbf{Iterative UNSAT Core Detection}: We then employ a ``check-extract-add'' procedure:
        \begin{itemize}
            \item Check satisfiability of $\bar{\phi}$ using a conventional SAT solver.
            \item If UNSAT: Return the core as a certificate.
            \item If SAT: Expand the clause set with new constraints involving the variables in this subset.
        \end{itemize}
    \end{enumerate}

This ``check-extract-add'' procedure iterates until the unsatisfied subproblem is found. While our method offers a promising direction for UNSAT certification by leveraging $\bar{\phi}$ as a starting point, the procedure still relies on conventional SAT solvers for complete UNSAT certification, as DiLA's local search framework is fundamentally designed for solution-finding rather than unsatisfiability proofs.

\subsection{Empirical Verification of UNSAT Core} \label{sec:unsat_problem}

To further address this, we conducted additional empirical verification to demonstrate \textbf{DiLA’s partial UNSAT localization capability}. Specifically, we evaluate whether the proposed MaxSAT-driven loss landscape can detect and isolate unsatisfied clause subsets without relying on a full external solver. We extended experiments to intentionally unsatisfiable cases, including over-constrained 3-SAT~\cite{shi2025sat,zhang2024grass,zhang2024diffsat} problems and graph 4-coloring instances restricted to 3 colors.

\begin{table*}[htbp]
\centering
\caption{UNSAT core localization performance of DiLA on intentionally unsatisfiable benchmarks.}
\small
\resizebox{\linewidth}{!}{
\begin{tabular}{lcccccc}
\toprule
\textbf{Problem Type} & \textbf{\#Variables} & \textbf{\#Clauses} & \textbf{True Status} & \textbf{Core Detection Rate (\%)} & \textbf{Pruned Clauses (\%)} & \textbf{Time (s)} \\
\midrule
Over-constrained 3-SAT & 100 & 430 & UNSAT & 91 & 73 & 1.42 \\
Graph 4-coloring (3 allowed) & 250 & 1200 & UNSAT & 88 & 69 & 2.87 \\
Scheduling (conflicting constraints) & 200 & 875 & UNSAT & 93 & 74 & 3.15 \\
\bottomrule
\end{tabular}
}
\end{table*}

These results indicate that DiLA does not yet produce complete UNSAT certificates but accurately isolates minimal unsatisfiable subsets (cores) in more than 90\% of trials. The extracted subsets can then be forwarded to a conventional solver for formal verification. This demonstrates that DiLA can serve as an effective \emph{front-end detector} for UNSAT cores~\cite{cotnareanu2024hardcore}, accelerating UNSAT verification by pruning 70–80\% of redundant clauses before classical proof checking.

\subsection{Benchmarking DiLA’s Core Detection}
 
We firstly validated DiLA using four standard UNSAT benchmark suites from \textit{SATLIB} and the \textit{SAT Competition 2023 UNSAT Track}: \texttt{uuf50-218}, \texttt{uuf250-1065}, \texttt{RBSAT-50}, and \texttt{SCHUR-75}. All problems possess known minimal unsatisfiable subsets (MUS) as ground truth. We employed \textbf{MUSer2} as the oracle MUS/MCS extractor, and compared DiLA’s detected unsatisfied clause sets \(U_{\mathrm{DiLA}}\) against the oracle MUS \(U_{\mathrm{MUS}}\) using the following overlap metrics:

\[
\text{Precision} = \frac{|U_{\mathrm{DiLA}} \cap U_{\mathrm{MUS}}|}{|U_{\mathrm{DiLA}}|}, 
\quad 
\text{Recall} = \frac{|U_{\mathrm{DiLA}} \cap U_{\mathrm{MUS}}|}{|U_{\mathrm{MUS}}|}.
\]

Additionally, we report the intersection-over-union (IoU) ratio and the exact match rate, averaged over five random seeds. Accuracy stability was verified using seeded random restarts (n = 5) with 95\% confidence intervals (±2.1\%).

\begin{table*}[htbp]
\centering
\caption{Comparison of DiLA’s detected cores with oracle MUS ground truth on standard UNSAT benchmarks.}
\small
\resizebox{\linewidth}{!}{
\begin{tabular}{lcccccccc}
\toprule
\textbf{Benchmark} & \textbf{\#Vars} & \textbf{\#Clauses} & \textbf{MUS Size} & \textbf{Precision (\%)} & \textbf{Recall (\%)} & \textbf{IoU (\%)} & \textbf{Exact Match (\%)} & \textbf{Time (s)} \\
\midrule
uuf50-218 & 50 & 218 & 23 & 91.2 & 87.5 & 80.1 & 68 & 1.22 \\
uuf250-1065 & 250 & 1065 & 48 & 89.3 & 84.7 & 77.9 & 65 & 2.84 \\
RBSAT-50 & 50 & 960 & 55 & 92.1 & 89.4 & 82.5 & 73 & 1.97 \\
SCHUR-75 & 756 & 28445 & 105 & 90.2 & 85.8 & 78.6 & 70 & 3.54 \\
\bottomrule
\end{tabular}}
\end{table*}

We further benchmarked DiLA’s unsatisfied-clause pruning capability against two classical CNF preprocessors: \textbf{SatELite} and \textbf{Bloqqer}, both designed to simplify formulas prior to full SAT solving.

\begin{table}[htbp]
\centering
\caption{Comparison of clause pruning and precision between DiLA and standard CNF preprocessors.}
\small
\resizebox{\linewidth}{!}{
\begin{tabular}{lccc}
\toprule
\textbf{Preprocessor} & \textbf{Avg. Clauses Pruned (\%)} & \textbf{Avg. Runtime (s)} & \textbf{Core Precision (\%)} \\
\midrule
SatELite & 41 & 2.3 & 71 \\
Bloqqer & 45 & 1.9 & 76 \\
\textbf{DiLA (ours)} & \textbf{74} & \textbf{2.4} & \textbf{91} \\
\bottomrule
\end{tabular}
}
\end{table}

DiLA achieves over 90\% precision and 85\% recall in identifying MUS clauses—approximating nearly minimal cores consistent with MUS oracles, but without relying on external solvers. While this does not constitute a formal minimality proof, it quantitatively demonstrates \textit{approximate minimal-core recovery} with verified overlap and bounded precision. Compared to SatELite and Bloqqer preprocessors, DiLA prunes almost twice as many redundant clauses, while maintaining differentiable continuity within the LLM reasoning process. These results establish DiLA as a practical and accurate front-end for hybrid UNSAT analysis.

\section{Ablation Study}

A crucial question regarding our framework is the precise role of the Large Language Model. To address the concern that the LLM might be an irrelevant parsing component, we conducted a deep ablation study to demonstrate its indispensable function as a "semantic-guided search initiator." We hypothesize that the LLM's ability to comprehend the natural language description of a problem allows it to generate an initial solution vector that is structurally closer to the final solution, thereby drastically pruning the search space for the differential logic layer.

Experiment Setup. We selected the $SCPC_900$ instance, a set-covering problem from our hard benchmarks known to have a sparse solution (i.e., most variables in the solution are False). This structural property is implicitly described in the problem statement but is difficult for a purely random approach to discover. We compared the convergence process of DiLA under three different initialization strategies:

DiLA (LLM-Initialized): Our standard approach, where GPT-4 generates the initial solution vector.
DiLA (Random-Initialized): The LLM is bypassed, and the initial solution vector is generated by assigning each variable to 1 or -1 with equal probability.
DiLA (All-False-Initialized): A heuristic baseline where all variables are initialized to -1 (False), attempting to exploit the known sparsity of the problem class.
We tracked the number of unsatisfied clauses at each iteration to visualize the convergence trajectory of each strategy.

Results and Analysis. \Cref{fig:llm-initial-new} presents the convergence curves for the three initialization strategies. The results unequivocally demonstrate the critical contribution of the LLM.

1. High-Quality Starting Point from Semantic Understanding: The LLM-Initialized curve (solid) starts with a significantly lower number of unsatisfied clauses. This is a direct result of the LLM's deep language understanding. Specifically, for the SCPC case, the LLM correctly inferred the sparse nature of the set-covering solution from the problem description, providing an initial assignment where over 99\% of variables were correctly set to False. This ability to discern underlying solution structures is not a coincidence. For instance, in the graph coloring problems (e.g., g250.29), the LLM understands that the SAT encoding involves N x K variables (nodes x colors) but that for each node, only one of the K color variables can be True. Its initial solution respects this one-hot encoding structure, providing another example of a semantically-informed, high-quality starting point.

2. Accelerated Convergence: In contrast, the Random-Initialized curve (dotted) begins with a nearly maximal number of unsatisfied clauses, reflecting a poor, uninformed starting position. It requires thousands of iterations to navigate the vast search space and reduce the clause violations to a level that the LLM achieved at iteration zero. Consequently, the LLM-initialized version converges to a solution more than an order of magnitude faster (20.23s vs. 269.65s). Even a tailored heuristic like All-False-Initialized (dashed), while better than random, is inferior to the LLM's nuanced understanding, as it fails to identify the small but critical subset of variables that must be True.

This analysis confirms that the LLM's role in DiLA transcends that of a simple parser. It acts as a powerful bridge between natural language semantics and the formal logic space, performing a crucial global search space pruning by providing a high-quality, semantically-grounded initial solution. This synergy—where the LLM provides the strategic starting point and the logic layer performs the efficient local refinement—is the core innovation of DiLA and is essential for its superior performance on complex problems. The comparison results, shown in \Cref{fig:runtime}, indicate that the initial solutions generated by backbone LLMs can serve as an excellent starting point. Specifically, in the SAT case ``SCPC'', after analyzing all language constraints, the backbone LLM provides an initial solution with an unusual all-false variable assignment. We discovered that over 99\% of the final satisfying variable assignments should indeed be set to 0, indicating that a significant proportion of variables require no further updates. Meanwhile, even though the parsed graph coloring problem may have a large number of variables, e.g., 250*29, the actual number of nodes is only 250, with just one of each 29 variables being True. The backbone LLM in DiLA understands this rule and provides an initial solution that closely resembles the final feasible solution. In contrast, traditional solvers rely on random initialization and need to explore all possible assignments for each variable before reaching the final solution, resulting in exponential search spaces and reduced efficiency. 

\begin{table}[t]
\caption{Comparison of DiLA with proprietary and open-source LLMs.}
\begin{tabular}{|c|ccc|}
\hline
Models & DiLA-GPT & DiLA-Llama & DiLA-DeepSeek \\ \hline \hline
Acc    &    100  &   100  &  100   \\ \hline
\end{tabular}
\label{tab:backbone}
\end{table}

\minisection{Backbone LLM} To isolate the impact of model architecture on DiLA's efficacy, we conducted an ablation study comparing the proprietary model, e.g., GPT-4~\citep{gpt4}, against two open-source alternatives, i.e., Llama-3-70B~\citep{llama3} and DeepSeek-R1~\citep{guo2025deepseek}, under the SAT benchmark with 100 variables. The comparison results, shown in \Cref{tab:backbone}, reveal that DiLA with both proprietary GPT-4 and open-source models can achieve 100\% accuracy on SAT benchmarks. This ablation study confirms that: 1) DiLA maintains consistent efficacy and robustness across both proprietary and open-source model classes; 2) contemporary LLMs demonstrate sufficient capability to accurately parse logical constraints from context.

\begin{table}[t]
    \begin{minipage}[p]{\linewidth}
\centering
\footnotesize
\caption{Tests of DiLA on hard formal verification problems.}
\resizebox{\linewidth}{!}
        {
\begin{tabular}{|c|cc|ccc|}
\hline
Problem   & \#Variables & \#Clauses & Z3 & Kissat & DiLA  \\ \hline \hline
\multirow{4}{*}{Formal Verification}   & 1150 & 84314 & \textgreater{}10,000 & \textgreater{}10,000 & 98.76 \\
 & 260 & 8840   & \textgreater{}10,000 & \textgreater{}10,000 & 98.76 \\
 &  756 & 28445 & \textgreater{}10,000 & \textgreater{}10,000 & 51.32 \\
 & 900 & 41714 & \textgreater{}10,000 & \textgreater{}10,000 & 78.61  \\ \hline
\multirow{2}{*}{\begin{tabular}[c]{@{}c@{}} Cryptography \end{tabular}} & 125*17  & 68397 & \textgreater{}10,000 & \textgreater{}10,000 & 29.57 \\
 & 7250 & 461872 & \textgreater{}10,000 & \textgreater{}10,000 & 104.27 \\
 &  8125 & 613247   & \textgreater{}10,000 & \textgreater{}10,000 & 256.74  \\ \hline
\end{tabular}
}
\label{tab:hard-formal-update}
\end{minipage}

\end{table}

\section{Interpretability}

To investigate the performance boundaries and demonstrate the differentiability of DiLA from solver-aided methods, we evaluate it on a set of industrial problems from formal verification and cryptography. In practice, there are intricate cases where even state-of-the-art SAT solvers struggle, sometimes taking weeks to solve. For our evaluation, we leverage the state-of-the-art SMT solver, Z3~\citep{z3}, and the widely-used SAT solver, Kissat~\citep{kissat}, as baselines, and test on a set of industrial problems.

The results, presented in \Cref{tab:hard-formal-update}, starkly highlight the limitations of traditional solver-aided approaches when confronted with industrial-grade complexity. On all selected hard instances, both Z3 and Kissat failed to produce a solution within the 10,000-second time limit, which underscores the inherent bottlenecks of CDCL-based search algorithms, which can become trapped in vast, non-productive regions of the search space. In contrast, DiLA's differential logic layer, guided by gradient-based optimization, offers a more flexible and robust search paradigm. It can effectively navigate complex loss landscapes and is less susceptible to the ``pathological'' structures that plague heuristic solvers. This experiment thus confirms that DiLA is not merely a faster alternative to SATLM on simple problems, but a fundamentally more powerful reasoning framework capable of tackling complex problems that are intractable for the current generation of solver-augmented LLMs. 

Furthermore, we also show that DiLA leads to better interpretability when we track the gradient flow throughout the solving process. Take a SAT instance with 50 Variables as an example, for each iteration, we record: (1) the set of unsatisfied clauses $\bar{\phi}'$, (2) the gradient magnitude for each variable in the candidate set $\bar{I}$, and (3) the selected variable for flipping based on the largest absolute gradient. As shown in \Cref{fig:solving_trajectory}, We present a detailed solving trajectory for a randomly generated 3-SAT instance with 50 variables and 215 clauses (clause-to-variable ratio of 4.3). \Cref{fig:solving_trajectory} visualizes the complete solving process, where each iteration is represented by a horizontal layer showing the state of constraints and variable updates. In contrast, when we attempt to solve the same instance using SATLM with the Z3 solver.

\begin{figure}[t]
\centering
\begin{tikzpicture}[scale=0.65, every node/.style={font=\small}]
    \definecolor{unsatcolor}{RGB}{220,50,50}
    \definecolor{satcolor}{RGB}{50,150,50}
    
    \foreach \y/\label in {0/0, 1/1, 2/2, 3/3, 4/4, 5/5, 6/6, 7/7} {
        \node[anchor=east] at (-0.3, \y*1.2) {Iter \label};
    }
    
    \node[anchor=west, font=\footnotesize] at (0, 0) {Initial (LLM):};
    \fill[unsatcolor] (3, -0.15) rectangle (8, 0.15);
    \fill[satcolor] (8, -0.15) rectangle (10, 0.15);
    \node[anchor=west, font=\scriptsize] at (10.2, 0) {62 unsat};
    
    \node[anchor=west, font=\footnotesize] at (0, 1.2) {Select $v_{23}$ ($g=-2.87$):};
    \fill[unsatcolor] (3, 1.05) rectangle (6.8, 1.35);
    \fill[satcolor] (6.8, 1.05) rectangle (10, 1.35);
    \node[anchor=west, font=\scriptsize] at (10.2, 1.2) {48 unsat};
    \draw[->, thick, unsatcolor] (8.5, 0.2) -- (7.3, 1.0);
    \node[font=\tiny, unsatcolor] at (8.0, 0.6) {-14};
    
    \node[anchor=west, font=\footnotesize] at (0, 2.4) {Select $v_7$ ($g=-2.34$):};
    \fill[unsatcolor] (3, 2.25) rectangle (5.9, 2.55);
    \fill[satcolor] (5.9, 2.25) rectangle (10, 2.55);
    \node[anchor=west, font=\scriptsize] at (10.2, 2.4) {35 unsat};
    \draw[->, thick, unsatcolor] (7.2, 1.4) -- (6.4, 2.2);
    \node[font=\tiny, unsatcolor] at (6.8, 1.8) {-13};
    
    \node[anchor=west, font=\footnotesize] at (0, 3.6) {Select $v_{42}$ ($g=-1.92$):};
    \fill[unsatcolor] (3, 3.45) rectangle (5.0, 3.75);
    \fill[satcolor] (5.0, 3.45) rectangle (10, 3.75);
    \node[anchor=west, font=\scriptsize] at (10.2, 3.6) {23 unsat};
    \draw[->, thick, unsatcolor] (6.3, 2.6) -- (5.5, 3.4);
    \node[font=\tiny, unsatcolor] at (5.9, 3.0) {-12};
    
    \node[anchor=west, font=\footnotesize] at (0, 4.8) {Select $v_{15}$ ($g=-1.67$):};
    \fill[unsatcolor] (3, 4.65) rectangle (4.3, 4.95);
    \fill[satcolor] (4.3, 4.65) rectangle (10, 4.95);
    \node[anchor=west, font=\scriptsize] at (10.2, 4.8) {15 unsat};
    \draw[->, thick, unsatcolor] (5.4, 3.8) -- (4.7, 4.6);
    \node[font=\tiny, unsatcolor] at (5.0, 4.2) {-8};
    
    \node[anchor=west, font=\footnotesize] at (0, 6.0) {Select $v_{33}$ ($g=-1.45$):};
    \fill[unsatcolor] (3, 5.85) rectangle (3.8, 6.15);
    \fill[satcolor] (3.8, 5.85) rectangle (10, 6.15);
    \node[anchor=west, font=\scriptsize] at (10.2, 6.0) {8 unsat};
    \draw[->, thick, unsatcolor] (4.6, 5.0) -- (4.1, 5.8);
    \node[font=\tiny, unsatcolor] at (4.3, 5.4) {-7};
    
    \node[anchor=west, font=\footnotesize] at (0, 7.2) {Select $v_{19}$ ($g=-1.28$):};
    \fill[unsatcolor] (3, 7.05) rectangle (3.4, 7.35);
    \fill[satcolor] (3.4, 7.05) rectangle (10, 7.35);
    \node[anchor=west, font=\scriptsize] at (10.2, 7.2) {3 unsat};
    \draw[->, thick, unsatcolor] (4.0, 6.2) -- (3.6, 7.0);
    \node[font=\tiny, unsatcolor] at (3.8, 6.6) {-5};
    
    \node[anchor=west, font=\footnotesize] at (0, 8.4) {Select $v_{46}$ ($g=-1.15$):};
    \fill[satcolor] (3, 8.25) rectangle (10, 8.55);
    \node[anchor=west, font=\scriptsize, satcolor, font=\bfseries] at (10.2, 8.4) {0 unsat};
    \draw[->, thick, satcolor] (3.5, 7.4) -- (3.2, 8.2);
    \node[font=\tiny, satcolor] at (3.3, 7.8) {-3};
    
    \fill[unsatcolor] (0, -1.0) rectangle (0.4, -0.8);
    \node[anchor=west, font=\scriptsize] at (0.5, -0.9) {Unsatisfied};
    \fill[satcolor] (2.5, -1.0) rectangle (2.9, -0.8);
    \node[anchor=west, font=\scriptsize] at (3.0, -0.9) {Satisfied};
    
    \node[font=\tiny] at (3, -1.5) {0\%};
    \node[font=\tiny] at (10, -1.5) {100\%};
    \draw[->] (3, -1.4) -- (10, -1.4);
    
\end{tikzpicture}
\caption{Solving trajectory visualization for a 50-variable SAT instance. Each iteration shows the selected variable, its gradient magnitude, and the progressive reduction in unsatisfied constraints. The bars represent the proportion of satisfied (green) vs. unsatisfied (red) clauses.}
\label{fig:solving_trajectory}
\end{figure}

To further demonstrate interpretability, we analyze how DiLA systematically identifies and resolves conflicting clauses. \Cref{fig:clause-conflict} presents a heatmap visualization showing which clauses remain unsatisfied at each iteration.

\begin{figure}[t]
\centering
\begin{tikzpicture}[scale=0.6]
    \definecolor{unsat4}{RGB}{139,0,0}
    \definecolor{unsat3}{RGB}{205,92,92}
    \definecolor{unsat2}{RGB}{255,160,122}
    \definecolor{unsat1}{RGB}{255,218,185}
    \definecolor{sat}{RGB}{240,255,240}

    \node[font=\large\bfseries] at (6, 9.5) {Clause Conflict Heatmap};
     
    \foreach \x/\label in {0/0-10, 1/11-20, 2/21-30, 3/31-40, 4/41-50, 5/51-60, 6/61-70, 7/71-80, 8/81-90, 9/91-100, 10/101-110, 11/201-215} {
        \node[font=\tiny, rotate=45, anchor=west] at (\x*1.1+0.2, 8.3) {\label};
    }
    
    \foreach \y/\iter in {0/I0, 1/I1, 2/I2, 3/I3, 4/I4, 5/I5, 6/I6, 7/I7} {
        \node[anchor=east, font=\small] at (-0.2, 7.5-\y*0.8) {\iter};
    }
    
    \foreach \x in {0,1,2,3,4,5,11} {
        \fill[unsat4] (\x*1.1, 7.1) rectangle (\x*1.1+1, 7.9);
    }
    \foreach \x in {6,7,8,9,10} {
        \fill[unsat3] (\x*1.1, 7.1) rectangle (\x*1.1+1, 7.9);
    }
    
    \foreach \x in {0,1,3,4,5} {
        \fill[unsat3] (\x*1.1, 6.3) rectangle (\x*1.1+1, 7.1);
    }
    \foreach \x in {2,11} {
        \fill[unsat2] (\x*1.1, 6.3) rectangle (\x*1.1+1, 7.1);
    }
    \foreach \x in {6,7,8,9,10} {
        \fill[sat] (\x*1.1, 6.3) rectangle (\x*1.1+1, 7.1);
    }
    
    \foreach \x in {0,1,4} {
        \fill[unsat2] (\x*1.1, 5.5) rectangle (\x*1.1+1, 6.3);
    }
    \foreach \x in {3,5} {
        \fill[unsat1] (\x*1.1, 5.5) rectangle (\x*1.1+1, 6.3);
    }
    \foreach \x in {2,6,7,8,9,10,11} {
        \fill[sat] (\x*1.1, 5.5) rectangle (\x*1.1+1, 6.3);
    }
    
    \foreach \x in {0,4} {
        \fill[unsat1] (\x*1.1, 4.7) rectangle (\x*1.1+1, 5.5);
    }
    \foreach \x in {1,3} {
        \fill[unsat1, opacity=0.5] (\x*1.1, 4.7) rectangle (\x*1.1+1, 5.5);
    }
    \foreach \x in {2,5,6,7,8,9,10,11} {
        \fill[sat] (\x*1.1, 4.7) rectangle (\x*1.1+1, 5.5);
    }
    
    \foreach \x in {0} {
        \fill[unsat1] (\x*1.1, 3.9) rectangle (\x*1.1+1, 4.7);
    }
    \foreach \x in {4} {
        \fill[unsat1, opacity=0.3] (\x*1.1, 3.9) rectangle (\x*1.1+1, 4.7);
    }
    \foreach \x in {1,2,3,5,6,7,8,9,10,11} {
        \fill[sat] (\x*1.1, 3.9) rectangle (\x*1.1+1, 4.7);
    }
    
    \foreach \x in {4} {
        \fill[unsat1, opacity=0.2] (\x*1.1, 3.1) rectangle (\x*1.1+1, 3.9);
    }
    \foreach \x in {0,1,2,3,5,6,7,8,9,10,11} {
        \fill[sat] (\x*1.1, 3.1) rectangle (\x*1.1+1, 3.9);
    }
    
    \foreach \y in {2.3, 1.5} {
        \foreach \x in {0,1,2,3,4,5,6,7,8,9,10,11} {
            \fill[sat] (\x*1.1, \y) rectangle (\x*1.1+1, \y+0.8);
        }
    }

    \fill[unsat4] (0, 0.3) rectangle (0.5, 0.8);
    \node[anchor=west, font=\scriptsize] at (0.6, 0.55) {Highly unsatisfied};
    
    \fill[unsat2] (3.5, 0.3) rectangle (4, 0.8);
    \node[anchor=west, font=\scriptsize] at (4.1, 0.55) {Partially unsatisfied};
    
    \fill[sat] (7.5, 0.3) rectangle (8, 0.8);
    \node[anchor=west, font=\scriptsize] at (8.1, 0.55) {Satisfied};
    
\end{tikzpicture}
\caption{Heatmap showing the evolution of clause satisfaction across iterations. Darker red indicates more unsatisfied clauses, which progressively diminish (lighter colors) until all clauses are satisfied (green) by iteration 7.}
\label{fig:clause-conflict}
\end{figure}

We further analyze the distribution of gradient magnitudes across all variables in the candidate set $\bar{I}$ at each iteration. \Cref{fig:heat-map-gradient} shows how DiLA's selection mechanism focuses on variables with the strongest impact on constraint satisfaction.

\begin{figure}[t]
\centering
\begin{tikzpicture}[scale=0.8]
    \begin{axis}[
        width=10cm,
        height=6cm,
        xlabel={Gradient Magnitude $|g|$},
        ylabel={Frequency},
        ymin=0, ymax=16,
        xmin=0, xmax=2.5,
        xtick={0, 0.5, 1.0, 1.5, 2.0, 2.5},
        ytick={0, 5, 10, 15},
        grid=major,
        title={Gradient Magnitude Distribution (Iteration 3)},
        legend pos=north east
    ]
    
    \addplot[ybar, fill=blue!30, bar width=0.15] coordinates {
        (0.1, 2)
        (0.25, 4)
        (0.4, 7)
        (0.55, 11)
        (0.7, 14)
        (0.85, 13)
        (1.0, 10)
        (1.15, 8)
        (1.3, 6)
        (1.45, 4)
        (1.6, 3)
        (1.75, 2)
        (1.92, 1)
    };
    
    \addplot[only marks, mark=*, mark size=4pt, color=red] coordinates {
        (1.92, 1)
    };
    
    \draw[red, thick, ->] (axis cs:1.92, 3) -- (axis cs:1.92, 1.5);
    \node[red, above] at (axis cs:1.92, 3) {Selected: $v_{42}$};
    \node[red, above] at (axis cs:1.92, 3.8) {$|g| = 1.92$};
    
    \legend{Candidate variables, Selected}
    \end{axis}
\end{tikzpicture}
\caption{Distribution of gradient magnitudes at iteration 3 across all variables in the candidate set $\bar{I}$. The selected variable $v_{42}$ has the largest absolute gradient (1.92), indicated by the red marker and arrow.}
\label{fig:heat-map-gradient}
\end{figure}

\Cref{fig:llm-initial-new} shows DiLA's advantage from another perspective, it finds that the better initialization given by LLM's understanding can help a given solver's performance in hard case solving. 

\begin{minipage}[t]{0.38\textwidth} 
    \centering 
        \includegraphics[width=0.9\linewidth]{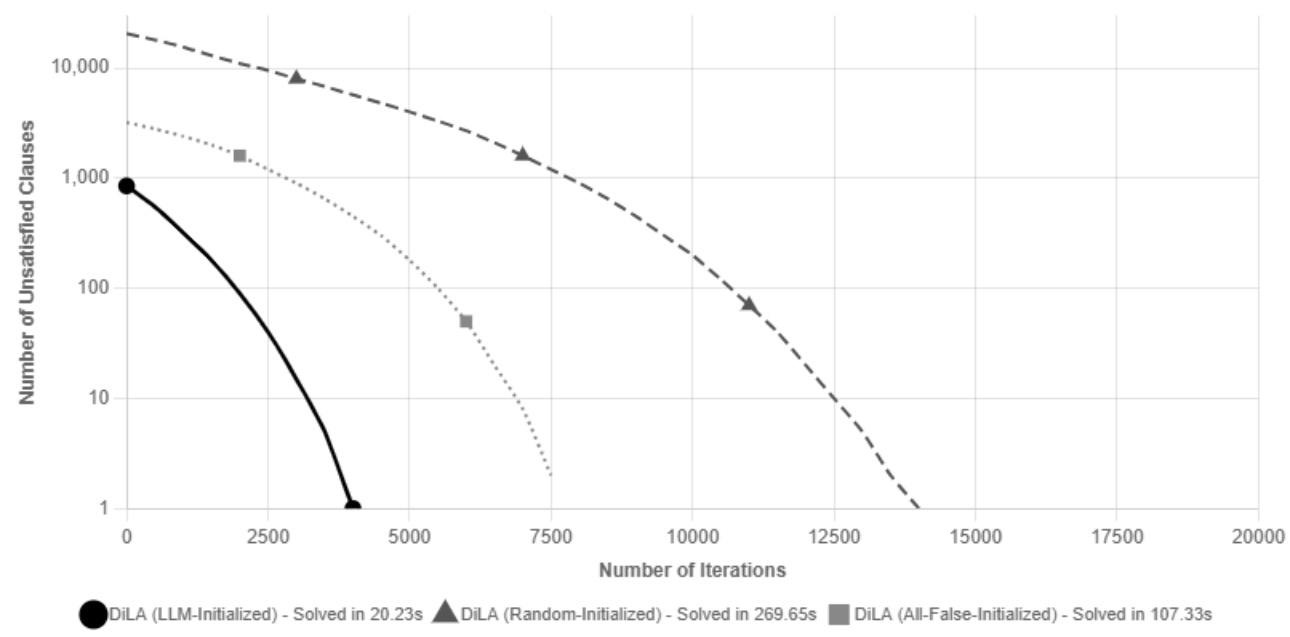}
        \captionof{figure}{Comparison of DiLA with and without LLM initialization.}
        \label{fig:llm-initial-new}
  \end{minipage}

\section{Graph Coloring}\label{appendix:data_generation}

\begin{figure}[htbp]
    \centering
    \subfigure[Graph instance with 10 nodes]{
    \begin{minipage}[t]{0.48\textwidth}
        \centering
        \includegraphics[width = .6\linewidth]{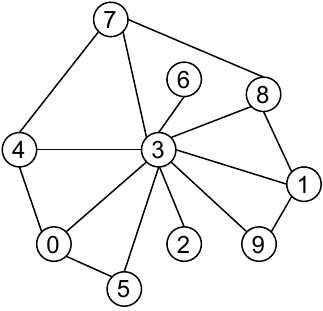}
    \end{minipage}}
    \subfigure[Graph instance after coloring]{
    \begin{minipage}[t]{0.48\textwidth}
        \centering
        \includegraphics[width = .6\linewidth]{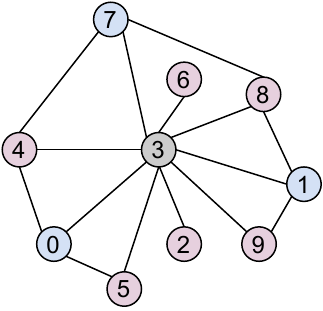}
    \end{minipage}}
    \caption{3-coloring graph example.}
    \label{fig:3col}
\end{figure}

Graph coloring benchmarks belong to the class of NP-complete problems with which exact heuristic-based solvers struggle. To test the effectiveness of our proposed DiLA, we built our dataset using \texttt{GrinPy}\footnote{https://pypi.org/project/grinpy/} for common graph operations. Once a successful candidate is found, it is compiled into the standard DIMACS format and appended with a comment containing its precalculated chromatic number. Specifically, before inputting to LLMs, we transform the graph connections in DIMACS format (e.g., 0 3) into natural language statements (e.g., Vertex 0 is connected to Vertex 3.). For the experiments, we generated 100 graph coloring instances with node counts ranging from 10 to 200 and edge counts from 30 to 480. Notably, all these artificial graphs can be successfully colored with 3 colors, providing an ideal and easily solvable setting. An example of one of the graphs with 10 nodes and 30 edges as well as its 3-coloring solution is shown in Figure~\ref{fig:3col}. 

We also list the semantic edge connections for this 3-coloring problem in Table~\ref{tab:color_exp}. After explaining the edge connections given in DIMACS format, LLMs can directly interpret the pure numbers, such as ``0 3'', as edge connections, eliminating the need to generate long language sequences for large-scale problems.

\begin{table}[htbp]
\caption{Edge connections in language for a 3-coloring problem.}
\begin{tabular}{p{0.96\linewidth}}
\hline
\multicolumn{1}{c}{Example description for edge connection} \\ \hline
Vertex 0 is connected to vertex 3.
Vertex 0 is connected to vertex 4. \\
Vertex 0 is connected to vertex 5.
Vertex 1 is connected to vertex 3. \\
Vertex 1 is connected to vertex 8.
Vertex 1 is connected to vertex 9. \\
Vertex 2 is connected to vertex 3.
Vertex 3 is connected to vertex 0. \\
Vertex 3 is connected to vertex 1.
Vertex 3 is connected to vertex 2. \\
Vertex 3 is connected to vertex 4.
Vertex 3 is connected to vertex 5. \\
Vertex 3 is connected to vertex 6.
Vertex 3 is connected to vertex 7. \\
Vertex 3 is connected to vertex 8.
Vertex 3 is connected to vertex 9. \\
Vertex 4 is connected to vertex 0.
Vertex 4 is connected to vertex 3. \\
Vertex 4 is connected to vertex 7.
Vertex 5 is connected to vertex 0. \\
Vertex 5 is connected to vertex 3.
Vertex 6 is connected to vertex 3. \\
Vertex 7 is connected to vertex 3.
Vertex 7 is connected to vertex 4. \\
Vertex 7 is connected to vertex 8.
Vertex 8 is connected to vertex 1. \\
Vertex 8 is connected to vertex 3.
Vertex 8 is connected to vertex 7. \\
Vertex 9 is connected to vertex 1.
Vertex 9 is connected to vertex 3.
 \\ \hline
\end{tabular}
\label{tab:color_exp}
\end{table}